\pdfoutput=1

\documentclass[11pt]{article}
\usepackage{graphicx}
\usepackage[preprint]{acl}
\usepackage{float}
\usepackage{subcaption}
\usepackage{amsmath}
\usepackage{amssymb}
\usepackage{times}
\usepackage{latexsym}
\usepackage[normalem]{ulem}
\usepackage{algorithm}
\usepackage[T1]{fontenc}

\usepackage[utf8]{inputenc}

\usepackage{microtype}

\usepackage{inconsolata}

\usepackage{graphicx}

\usepackage{algpseudocode}
\usepackage{float}

\usepackage{xcolor}

\usepackage{amsmath,amssymb}
\usepackage{booktabs}            
\usepackage{multirow}            
\usepackage{array}               
\usepackage{xcolor}              
\usepackage{adjustbox}           
\usepackage{float}               
\usepackage{placeins}            
\usepackage{pgfplots} 

\title{FedSEAM: Federated Continual Learning of LLMs via Subspace Expert Agreement Merging}

\author{
\textbf{Fatema Siddika}\textsuperscript{1,2},
\textbf{Abhijit Chunduru}\textsuperscript{2,3},
\textbf{Zilinghan Li}\textsuperscript{2},
\textbf{Ali Jannesari}\textsuperscript{1},
\textbf{Ravi Madduri}\textsuperscript{2} \\
\textsuperscript{1}Iowa State University, Ames, USA \\
\textsuperscript{2}Argonne National Laboratory, USA \\
\textsuperscript{3}University of Massachusetts Amherst, Amherst, USA \\
\texttt{\{fatemask, jannesar\}@iastate.edu} \\
\texttt{\{schunduru,  zilinghan.li, madduri\}@anl.gov} \\
}

\usepackage{amsmath,amssymb,amsthm,mathtools}
\usepackage{booktabs,multirow,array}
\usepackage{algorithm}
\usepackage{algpseudocode}
\usepackage{pgfplots}
\pgfplotsset{compat=1.17}
\usepgfplotslibrary{polar}

\newcommand{\std}[1]{\textsubscript{\scriptsize$\pm$#1}}              

\newtheorem{theorem}{Theorem}
\newtheorem{proposition}{Proposition}
\newtheorem{lemma}{Lemma}

\newcommand{\method}{FedSEAM}
\newcommand{\SEAM}{SEAM}

\newcommand{\E}{\mathbb{E}}
\newcommand{\norm}[1]{\left\lVert #1 \right\rVert}
\newcommand{\fro}[1]{\lVert #1\rVert_{F}}
\newcommand{\Es}{E_{\mathrm{s}}}
\newcommand{\PiC}{\Pi_{\mathcal{C}}}

\newcommand{\PiCT}{\Pi_{\mathcal{C}\cap\mathcal{T}}}
\newcommand{\QC}{Q_{\mathcal{C}}}

\begin{document}
\maketitle
\begin{abstract}
Federated continual learning trains a model on a sequence of tasks across several clients without sharing their data while retaining knowledge of  earlier tasks. Sparse mixture-of-experts models support this setting by adapting selected parameters on a frozen backbone. However, clients learn different expert content as tasks evolve, making it difficult to share knowledge across clients without disrupting task-specific and client-specific specialization.
Matching tile coordinates alone does not establish whether their learned content is compatible. We present \method{}, a federated continual learning framework that organizes task knowledge into sparse experts and selectively shares their content across tasks and clients. On each client, task-aware tile selection allocates capacity to a new task, while a subspace-based compatibility test determines which existing expert tiles to reuse and where to introduce
fresh capacity. On the server, \textbf{S}ubspace \textbf{E}xpert \textbf{A}greement \textbf{M}erging (\SEAM{}) compares client updates as subspaces and moves each client towards consensus only along agreed-upon directions, leaving the rest of its update unchanged. We identify two remaining sources of forgetting, routing drift and merge erasure, and address them with three protections, one of which exactly preserves tile components in a selected prior-overlap subspace at every merge. We evaluate \method{} on TRACE using Qwen3-4B and Llama-3.2-3B, with component-wise ablations and studies of data size to assess its balance between learning new tasks and retaining prior knowledge.
\end{abstract}

\section{Introduction}
Federated learning (FL) trains a shared model across clients that keep their raw data private and
send only model updates~\citep{mcmahan2017fedavg}, which suits sensitive or widely distributed data. For large
language models, FL has become a practical way to fine-tune a common model on data that never leaves
its owners~\citep{zhang2023fedit}. Language models are increasingly adapted after deployment by organizations that cannot share data, including hospitals, banks, and research laboratories. As their data and requirements evolve, these organizations must acquire new capabilities while retaining earlier knowledge, reflecting the stability--plasticity dilemma of continual learning~\citep{kirkpatrick2017ewc,lopezpaz2017gem}.

Federated continual learning (FCL)~\citep{yoon2021fedweit} studies this combined setting, which is
more challenging than either federated or continual learning alone. Client updates are computed on different distributions and often disagree in direction, and
when they are combined coordinate by coordinate, they can cancel or overwrite task-specific information. Each party also sees only its own data, so clients collaborate by exchanging parameter updates.

Most federated continual learning methods have been developed and evaluated on vision tasks. Some
decompose model parameters to transfer task knowledge between clients~\citep{yoon2021fedweit}, an idea
later applied to text classification~\citep{chaudhary2022fedseit}, while others counter forgetting with
generated replay or distillation~\citep{wuerkaixi2024affcl,ma2022cfed}. Recent work turns to
multimodal models, either combining regularization with replay~\citep{liu2026fedcmm} or keeping
knowledge in separate modules that are selected or fused for each input~\citep{guo2025fcit,guo2026fedduet},
an idea shared with mixtures of experts, which combine modules through a router~\citep{shazeer2017moe,dou2024loramoe}. Routing alone, however, does not prevent interference between tasks~\citep{wu2026prism}, and these methods leave open how clients' versions of the same
expert should be combined when task-specific knowledge is represented by sparse updates to a language model.

A recent design for continual learning of language models adapts a frozen backbone through sparse
experts~\citep{siddika2026seta}: each task trains a small set of weight tiles, and a
router combines the experts without task labels. Extending this design to the federated setting raises
three challenges.
First, clients trained on different data drive shared parameters in different directions, and
coordinate-wise aggregation methods such as federated averaging or task arithmetic~\citep{ilharco2023task}
dilute client-specific knowledge. Second, successive tasks tend to select the same tiles, because the same input channels dominate gradient magnitudes across tasks, and sharing tiles based on position alone cannot distinguish compatible tasks from conflicting ones. Third, freezing experts from completed tasks
is insufficient to prevent forgetting, since the router and shared experts continue to change.

\begin{figure}[htbp]
\centering
\includegraphics[width=\columnwidth]{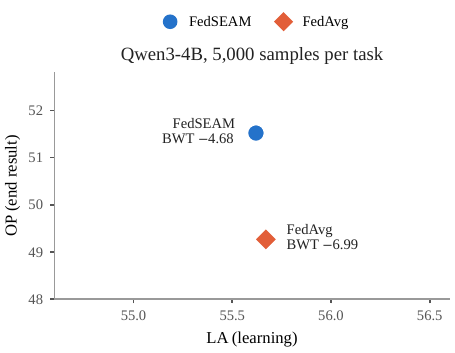}
\caption{LA and OP of \method{} and FedAvg on Qwen3-4B with 5,000 training
samples per task (client mean); labels give BWT. \method{} achieves comparable LA and an OP that is 2.5 points higher.}
\label{fig:scatter5000}
\end{figure}

To address these challenges, we propose \textbf{Fed}erated \textbf{S}ubspace \textbf{E}xpert \textbf{A}greement \textbf{M}erging (\method{}), a sparse mixture-of-experts framework for federated continual learning. On each client, task-aware tile selection and selective expert reuse allow new tasks to build on compatible knowledge while allocating fresh capacity where needed. The server then aggregates shared expert tiles through subspace agreement, combining compatible client contributions while retaining components outside the agreed subspace.
Figure~\ref{fig:scatter5000} previews the main result on Qwen3-4B with 1,000
samples per client: \method{} achieves learning accuracy comparable to FedAvg over tiles, with higher final performance and less forgetting in
the available evaluations.
Two research questions guide the paper.\\
\textbf{RQ1:} How can sparse expert content be shared across tasks and
aggregated across clients while retaining task and client specialization?\\
\textbf{RQ2:} What causes forgetting when earlier experts are frozen,
and how can expert training and aggregation limit it?
Our main contributions are as follows:\\
\textbf{Contribution 1:} \method{}, a sparse mixture-of-experts framework for federated continual learning, with task-aware tile selection and selective expert reuse.\\
\textbf{Contribution 2:} \SEAM{}, a personalized aggregation method that shares expert content along agreed update directions while retaining client-specific components outside them.\\
\textbf{Contribution 3:} An analysis of routing drift and merge erasure, with protections that include exact preservation of tile components in a selected prior-overlap subspace during merging.\\
\textbf{Contribution 4:} Evaluation on TRACE with Qwen3-4B and Llama-3.2-3B, examining learning, retention, and the effects of individual components.

\section{Related Work}\label{sec:related}

\textbf{Federated Continual Learning.}
FCL seeks to retain knowledge across tasks while transferring
relevant information between heterogeneous clients.
FedWeIT~\citep{yoon2021fedweit} separates global and sparse
task-specific parameters and learns weighted inter-client
transfer. FedSeIT~\citep{chaudhary2022fedseit} extends this
approach to text classification, using domain overlap to select
relevant historical tasks. Other methods retain knowledge through
generative replay~\citep{wuerkaixi2024affcl} or distillation
with unlabeled surrogate data~\citep{ma2022cfed}.
\method{} builds on selective parameter sharing, using expert
weights and task-start snapshots without replay or auxiliary
distillation data.

\textbf{Continual Adaptation of Language Models.}
O-LoRA~\citep{wang2023olora} limits interference through
orthogonal low-rank task subspaces, while
TreeLoRA~\citep{qi2025treelora} organizes layer-wise adapters
by gradient similarity. LoRAMoE~\citep{dou2024loramoe}
uses routed low-rank experts to mitigate world-knowledge
forgetting during instruction tuning. In federated settings,
DISCO~\citep{guo2025fcit} maintains a server-side cache of
task subspaces and selectively activates them at inference.
Fed-Duet~\citep{guo2026fedduet} combines shared semantic
prompts with personalized adapters. The closest approach to our client-side learner is SETA \citep{siddika2026seta}, which selects sparse expert blocks by gradient magnitude and shares blocks across tasks based on positional overlap in a centralized setting.
CRAFT~\citep{hossen2026craft} instead adapts hidden representations through
low-rank interventions with a forgetting-aware objective.
These methods guide the
organization and reuse of task knowledge. \method{} additionally
determines which directions within shared tiles receive a
consensus correction and which remain client-specific.

\section{Problem Formulation}\label{sec:problem}

\textbf{Setting.} $K$ clients learn $N$ tasks $T_1,\dots,T_N$ in a fixed order.
For task $T_t$, client $k$ holds a private shard $D_t^k$, and training proceeds for $R$ communication rounds. Raw data never leave the clients, and data from completed tasks are no longer available.

\textbf{Model.} The backbone is frozen. At each adapted layer with frozen weight
$W_0$, sparse expert updates are combined through sigmoid gates:
\begin{equation}
y = W_0 x + \sum_{e} \sigma\big(g_e^\top x + b_e\big)\, \Delta W_e\, x.
\label{eq:model}
\end{equation}
Each $\Delta W_e$ is nonzero only on a set of $l\times l$ blocks of $W_0$, called
\emph{tiles}, each with a fixed position (layer, row, column), and $(g_e,b_e)$ is
the expert's \emph{gate row}. We use $l{=}128$ on the value projections (36
layers of $8\times20$ tiles for Qwen3-4B). Gates are independent sigmoids, and a
new expert starts at $\Delta W_e{=}0$, so adding it leaves the output unchanged.
Each task trains its own expert and, for tiles it shares with earlier experts, a
shared expert $\Es$ that starts from the learned values of one earlier expert per layer;
experts of finished tasks are frozen. A task may train $B$ tiles, a fraction
$\beta{=}0.15$ of all tiles ($B{=}864$ for Qwen3-4B).

\textbf{Objective and metrics.} At task $t$, the clients solve
\begin{equation}
\min_{\Theta_t}\;\sum_{k} p_k\,
\E_{(x,y)\sim D_t^k}\big[-\log P(y\mid x;\Theta_t)\big]
+\mathcal{L}_{\mathrm{reg}}  
\label{eq:objective}
\end{equation}
over the parameters $\Theta_t$ trained at task $t$, using only local training and
server aggregation, with $p_k=1/K$ and $\mathcal{L}_{\mathrm{reg}}$ the anchoring
penalty of \S\ref{sec:training}. With $R_{i,j}$ the test score on $T_j$ after
learning $T_i$, averaged over the clients' final-round models, we report
learning accuracy (LA), overall performance (OP) and backward transfer (BWT):
\begin{equation}
\begin{aligned}
\mathrm{LA}&=\tfrac{1}{N}\textstyle\sum_{i}R_{i,i},\quad
\mathrm{OP}=\tfrac{1}{N}\textstyle\sum_{j}R_{N,j},\\
\mathrm{BWT}&=\tfrac{1}{N-1}\textstyle\sum_{j<N}\bigl(R_{N,j}-R_{j,j}\bigr),
\end{aligned}
\label{eq:metrics}
\end{equation}
where a negative BWT indicates forgetting.

\textbf{Forgetting with frozen experts.} Freezing experts from completed tasks does not
prevent forgetting. A frozen expert's gate changes when its gate row is trained
or its layer input shifts (\emph{routing drift}), and a shared expert loses
earlier content when the merge moves it along directions that earlier tasks
depend on (\emph{merge erasure}); Proposition~\ref{prop:decomp} makes both
precise.

\begin{figure*}[htbp]
\centering
\includegraphics[width=0.9\linewidth]{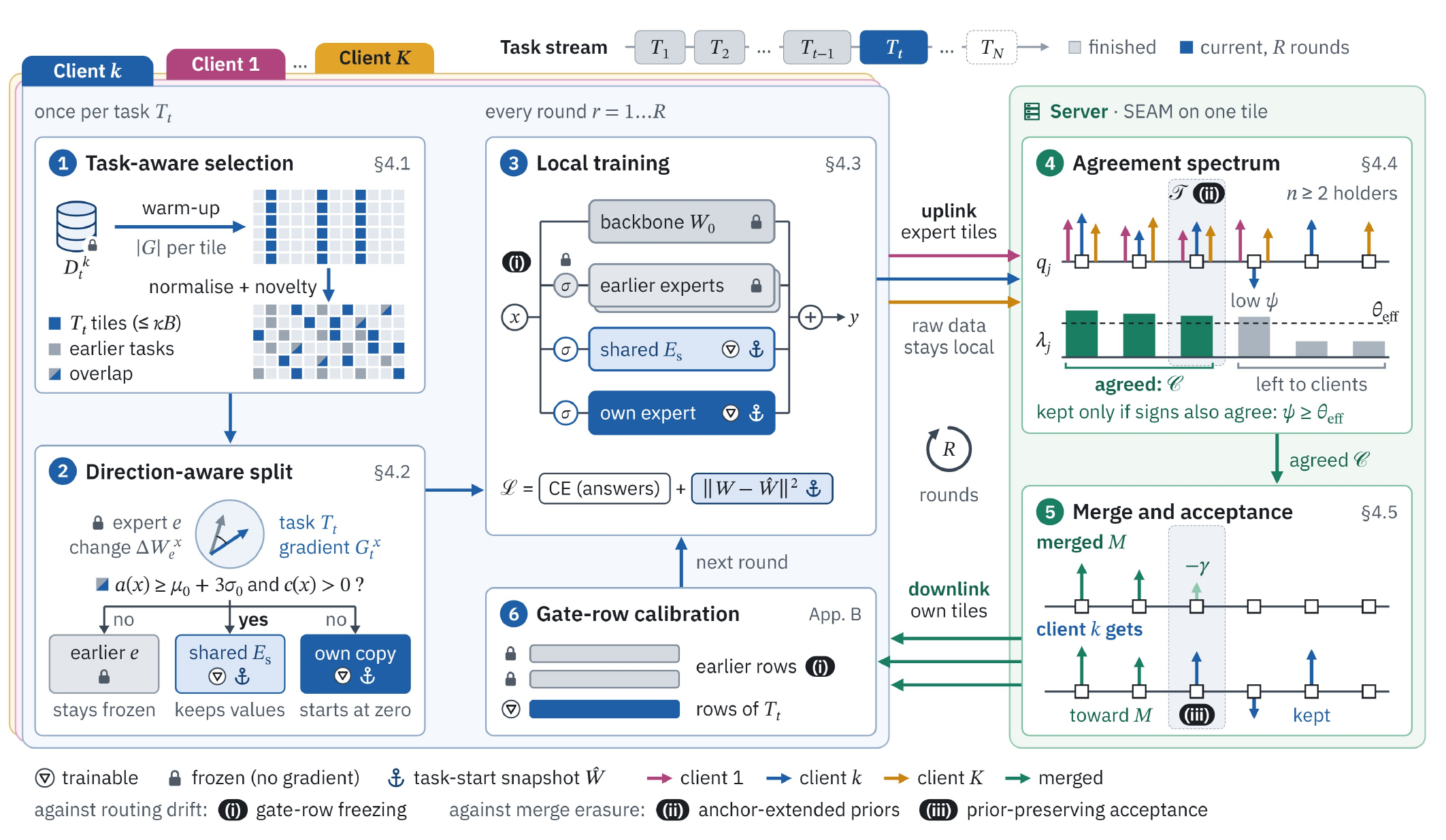}
\caption{Overview of \method{}. Once per task, each client
selects tiles for $T_t$ (1) and shares a tile with an earlier expert only if
both updates use the same input directions and agree in sign (2). In each
round, clients train the current experts (3); the server keeps the directions
$\mathcal{C}$ that most clients move along in the same sense (4) and merges
the tiles along them (5); each client then recalibrates its current gate
rows (6). (i)--(iii) mark the protections against routing drift and merge
erasure.}
\label{fig:framework}
\end{figure*}

\section{Methodology}\label{sec:method}
\method{} runs the same procedure for every task (Figure~\ref{fig:framework};
Algorithm~\ref{alg:fedseam}). 
\subsection{Warm-up and Task-aware Selection}\label{sec:selection}
The warm-up fine-tunes the whole backbone on the client's shard for at most one
pass, with an adaptive stopping rule. The client accumulates the gradient $G$
and then discards the fine-tuned model. Client $k$ scores the tile in layer
$\ell$, row $r$ and column $q$ of the tile grid by the mean absolute value of its accumulated gradient, $m^k(\ell,r,q)$,
as in SETA~\citep{siddika2026seta}. The score map is then processed in
three steps that use only local data and task history.

\textbf{Column correction.} Ranking tiles by raw gradient magnitude favours a few hidden channels whose
activations are large for every input, so whole columns of the tile grid score
high regardless of the task. In a preliminary run on Qwen3-4B with five clients,
the second task reused 763 of the first task's 864 tiles (88\%), against 15\%
for independent selections of the same size.

We first standardise scores within each tile column, then across the layer:
\begin{equation}
\tilde m^{k}(\ell,r,q)=\frac{m^{k}(\ell,r,q)-\mu^{k}_{\ell,q}}{\sigma^{k}_{\ell,q}+\varepsilon},
\label{eq:fixcol}
\end{equation}
followed by $z^{k}_\ell=\max\{0,\mathrm{std}(\tilde m^{k}_\ell)\}$, where $\mu^{k}_{\ell,q}$ and
$\sigma^{k}_{\ell,q}$ are taken over the rows of column $q$ and $\mathrm{std}$ standardises a layer. A tile receives a high score only if it stands out among the tiles associated with the same input channels, which removes this column effect (Proposition~\ref{prop:colinv}).

\textbf{Novelty.} From the second task on, with $z^{\,\mathrm{prev}}_\ell$ the mean map of the client's
earlier tasks after the same correction, the score is
\begin{equation}
s=\alpha_t\,z_\ell+(1-\alpha_t)\,\big|z_\ell-z^{\,\mathrm{prev}}_\ell\big|,
\label{eq:novelty}
\end{equation}
with $\alpha_t=1/(1+\ln t)$, where both maps are standardised per layer. The weight on novelty grows
with the number of tasks, because unused tiles become harder to find.

\textbf{Budget.} The client selects the $N_t$ highest-scoring tiles for the task, with $N_t$ set
by a walk down the ranked list (Algorithm~\ref{alg:walk}): layers less used by
earlier tasks get larger quotas, tiles used by two or more earlier tasks are
skipped, and after $B$ tiles the walk stops once the share of unused tiles
stabilizes, at most at $\kappa B$ ($\kappa{=}1$ in the main configuration); the quotas and the skip rule set only this count, not which tiles are selected, and with $\kappa{=}1$ the walk returns $N_t{=}B$ (Appendix~\ref{app:method}).
Clients select different tiles; \SEAM{} merges those that at least two clients
selected.

\subsection{Direction-Aware Expert Sharing}\label{sec:split}
When a selected tile $x$ already belongs to an earlier expert $e$, the client
either shares it, so the new task continues from the learned values, or keeps it
frozen in $e$ and gives the new task a fresh copy that starts at zero. Sharing by position alone, as in the Split-on-Share step of
SETA~\citep{siddika2026seta}, cannot tell whether the two tasks move the tile in
compatible directions, so we compare the earlier task's learned change $\Delta W_e^{x}$ with
the new task's warm-up gradient $G_t^{x}$.

\textbf{Direction test.} Let $V_{\xi,r}(A)$ hold the leading right singular
vectors of $A$ that carry a fraction $\xi$ of its energy, at most $r$ of them,
and let $V_e=V_{0.9,8}(\Delta W_e^{x})$ and $V_t=V_{0.9,8}(G_t^{x})$ have $r_e$
and $r_t$ columns. Their overlap is the mean squared cosine of the principal
angles,
\begin{equation}
a(x)=\fro{V_e^{\top}V_t}^{2}\big/\min\{r_e,r_t\}\in[0,1].
\label{eq:score}
\end{equation}
Because gradients within one network resemble one another, the threshold is
calibrated per tile: the test passes if
\begin{equation}
a(x)\ge\mu_0(x)+3\,\sigma_0(x),
\label{eq:null}
\end{equation}
where $\mu_0$ and $\sigma_0$ are the mean and standard deviation of the scores
of $V_e$ against the new task's gradients on tiles it selected in the same layer
(Appendix~\ref{app:method}).

\textbf{Sign test.} Because $A$ and $-A$ have the same right singular vectors,
$a(x)$ can equal $1$ for opposite updates. A shared tile must therefore also
satisfy
\begin{equation}
c(x)=\frac{\langle \Delta W_e^{x},-G_t^{x}\rangle}
{\fro{\Delta W_e^{x}}\,\fro{G_t^{x}}}>0.  
\label{eq:sign}
\end{equation}
This test measures alignment with the new task's accumulated
warm-up descent direction; it does not guarantee preservation
of the earlier task's loss.

\textbf{Size threshold and shared expert.} A layer is split only if at least
$T=\phi B$ tiles pass both tests ($\phi{=}0.015$, $T\approx13$ for Qwen3-4B) and
the earlier expert and the new task's own expert each keep none or at least $T$
tiles there, so no expert holds a small remainder; otherwise the new task trains
fresh copies and the earlier expert stays intact. Passing tiles form one shared
expert per task, and the task's other tiles form its own expert, which starts at
zero. In each layer, the shared expert copies the gate row and values of the
earlier expert that contributes the most tiles; the other earlier experts keep
their tiles, and the shared expert holds those coordinates as zero-initialized
copies. The split therefore leaves every layer's output unchanged
(Algorithm~\ref{alg:split}, Proposition~\ref{prop:splitpres}).

\subsection{Local Training}\label{sec:training}
Each client trains the experts of the current task and their gate rows by
minimizing
\begin{equation}
\begin{aligned}
\mathcal{L}={}&\frac{1}{|\mathcal{Y}|}\sum_{i\in\mathcal{Y}}-\log P(y_i\mid y_{<i},x)\\
&+\mu\sum_{e}\fro{\Delta W_e-\hat W_e}^{2},
\end{aligned}
\label{eq:loss}
\end{equation}
where $\mathcal{Y}$ is the set of answer tokens in the batch, $e$ runs over the
experts of the current task, and $\hat W_e$ is a snapshot of $\Delta W_e$ taken
once per task after the split (at the start of training for the first task),
with $\mu{=}0.1$. For a shared expert, the penalty anchors the inherited values;
for the task's own expert, whose snapshot is zero, it acts as weight decay. A
sample longer than the context window keeps its whole answer and loses the
middle of its prompt (Appendix~\ref{app:method}).

\subsection{\SEAM{}: Agreement Spectrum}\label{sec:spectrum}
\SEAM{} merges client updates only along input directions that most clients use
and move in the same sense; a direction that a single client uses stays private
to it.

\textbf{Agreement spectrum.} The server groups the received tiles by coordinate
(layer, row, column), so client updates are compared without any alignment step.
A bin holds the clients that selected that tile, and bins with fewer than two
holders stay local. For client $k$ in a bin, let $\Delta^k$ be its tile, $U^k$ its change since the start of the task, $U^k=\Delta^k-\hat W^k$ (Table~\ref{tab:abl_ref} measures it from the start of the round instead), and
$Q_k=V_{0.9,32}(U^k)$ the leading input directions of $U^k$. \SEAM{} forms the weighted mean projector
\begin{equation}
\widehat{P}=\sum_{k}\pi_k\,Q_kQ_k^{\top}=Q\,\mathrm{diag}(\lambda_j)\,Q^\top,
\label{eq:meanproj}
\end{equation}
with weights $\pi_k\propto\max\{0,\cos(U^k,\bar U)\}$ that sum to one and reduce
the influence of a client whose update opposes the mean update $\bar U$. The
eigenvalue $\lambda_j\in[0,1]$ is the weighted mean squared cosine between the
eigenvector $q_j$ and the client subspaces: $\lambda_j=1$ when every client's
subspace contains $q_j$, and $\lambda_j=0$ when none does. The spectrum is
smoothed over the rounds of a task (Appendix~\ref{app:method}). Direction $q_j$
is accepted if
$\lambda_j\ge\theta_{\mathrm{eff}}=\max\{\theta,\lambda_{\mathrm{chance}}+0.05\}$,
with $\theta{=}0.7$ and $\lambda_{\mathrm{chance}}$ the largest eigenvalue of the
same projector for one random subspace per client, each of the clients' average
rank. The accepted directions form the columns of $\QC$.

\textbf{Sign agreement.} Because $U$ and $-U$ have the same right singular
vectors, $\lambda_j$ cannot tell whether clients move along $q_j$ in the same or
in opposite senses. \SEAM{} therefore keeps only the accepted directions along
which the normalized updates also agree in sign. With
$A_k=(U^k/\norm{U^k})\,\QC$, $S=\sum_k A_k$ and $Z=\sum_k A_k^\top A_k$, the
signed agreements $\psi_j$ are the eigenvalues of
\begin{equation}
\Psi=\tfrac{1}{n}\,Z^{-1/2}S^{\top}S\,Z^{-1/2},
\label{eq:signagree}
\end{equation}
where $n$ is the number of clients in the bin. Along a direction, $\psi_j$
equals the fraction of clients that move identically when the others do not
move, and is close to zero when their movements cancel; because the updates are
normalized, it does not depend on their magnitudes. The directions with
$\psi_j\ge\theta_{\mathrm{eff}}$ span the common subspace $\mathcal{C}$. From
here on, $\QC$ denotes their basis, $\PiC=\QC\QC^{\top}$ its projector, and
$\lambda_j$ their weights recomputed from $\widehat{P}$ in this basis.

\textbf{Private directions.} Acceptance counts how many clients use a direction
rather than averaging their values. A direction that only client $k$ uses, and
that is orthogonal to the other clients' subspaces, receives
$\lambda_j\approx\pi_k$. This is below $\theta_{\mathrm{eff}}$ unless the other
clients carry almost no weight, so the direction stays with client $k$, whereas
a coordinate-wise average would scale it by the client's weight.

\subsection{Merged Tile and Acceptance}\label{sec:emit}
\textbf{Merged tile.} At each coordinate, the prior subspace $\mathcal{T}$ is the
span of $V_{0.9,32}$ of the tiles that the clients' earlier experts hold there,
stacked (computed from their Gram matrices), and $\mathcal{C}\cap\mathcal{T}$ is
spanned by the principal directions of $\mathcal{C}$ whose squared cosine with
$\mathcal{T}$ is at least $0.5$. The merged tile is the mean $\bar\Delta$ of the
clipped client tiles, restricted to $\mathcal{C}$ and reduced along earlier
tasks,
\begin{equation}
M=\bar\Delta\,\PiC,\qquad M\leftarrow M-\gamma\,M\,\PiCT,
\label{eq:merged}
\end{equation}
where $\gamma=\fro{M\PiCT}^2/\fro{M}^2$ is the share of the energy of $M$ in
$\mathcal{C}\cap\mathcal{T}$.

\textbf{Acceptance.} Each client's tile moves towards $M$ only along accepted
directions, by their agreement:
\begin{equation}
\Delta^k\leftarrow\Delta^k-(\Delta^k-M)\,W_\lambda,
\label{eq:accept}
\end{equation}
with $W_\lambda=\QC\,\mathrm{diag}(\lambda_j)\,\QC^{\top}$. Directions outside
$\mathcal{C}$, on which clients disagree, stay in the client's tile, so
client-specific knowledge is kept rather than averaged away
(Lemma~\ref{lem:preserve}a). Appendix~\ref{app:method} defines the bin labels
used in our analysis and an optional variant, used only as an ablation, that
moves the residual $\Delta^k(I-\PiC)$ into a private expert.

\subsection{Stability Protections}\label{sec:protect}
Three protections, each an independent switch, act against routing drift and
merge erasure.

\textbf{(i) Gate-row freezing.} During training and calibration, the gate rows
of experts of earlier tasks receive no gradient; since each sigmoid gate depends
only on its own row, these gates stay fixed for a fixed layer input. The gate
row of the current task's shared expert still trains, so the routing of its
inherited tiles can change.

\textbf{(ii) Anchor-extended priors.} The snapshots $\hat W$ of the current
experts, taken after the split, are added to the tiles from which $\mathcal{T}$
is computed, so $\mathcal{T}$ also covers the tiles a shared expert inherited.

\textbf{(iii) Prior-preserving acceptance.} The pull excludes
$\mathcal{C}\cap\mathcal{T}$,
\begin{equation}
\Delta^k\leftarrow\Delta^k-(\Delta^k-M)\,W_\lambda\,(I-\PiCT),
\label{eq:protacc}
\end{equation}
which leaves the tile's component in $\mathcal{C}\cap\mathcal{T}$ unchanged
(Lemma~\ref{lem:preserve}b). The rest of $\mathcal{C}$ can still change the
tile's component in $\mathcal{T}$, so the protection covers
$\mathcal{C}\cap\mathcal{T}$, not all of $\mathcal{T}$.

When the consensus opposes what was learned, that is, when
$\cos(M\PiCT,\bar{\hat W}\PiCT)<-0.1$ with $\bar{\hat W}$ the mean snapshot at
that coordinate, $\gamma$ is set to $1$, which removes the component of $M$ in
$\mathcal{C}\cap\mathcal{T}$. Because (iii) already excludes this component from the pull, the test changes client tiles only through the unequal weights in $W_\lambda$ (Appendix~\ref{app:method}); its main effect is on the server model, which we do not evaluate (Appendix~\ref{app:efficiency}).

\section{Theoretical Analysis}\label{sec:theory}
Proofs and further propositions are in Appendix~\ref{app:proofs}.

\begin{proposition}[Forgetting decomposition]\label{prop:decomp}
For an input $x$ of an earlier task $j$, held fixed at the input of this layer, and a checkpoint $t'>j$,
the output change of the layer is
\begin{align}
\Delta y={}&\textstyle\sum_e \Delta\sigma_e(x)\,\Delta W_e^{(j)}x \nonumber\\
&+\textstyle\sum_e \sigma_e^{(t')}(x)\big(\Delta W_e^{(t')}-\Delta W_e^{(j)}\big)x \nonumber\\
&+\textstyle\sum_{e\,\mathrm{new}}\sigma_e^{(t')}(x)\,\Delta W_e^{(t')}x ,
\end{align}
where $e$ ranges over the experts present at the end of task $j$ (an expert removed by a later split
counts with $\Delta W_e^{(t')}=0$ and $\sigma_e^{(t')}=\sigma_e^{(j)}$) and ``new'' denotes experts
created after task $j$; that is, routing drift, weight change and intrusion of new experts.
\end{proposition}
For a fixed layer input, protection (i) removes the first term, and a split
leaves $\Delta y$ unchanged. The remaining change comes from weight drift, which
the penalty in Equation~\eqref{eq:loss} and protections (ii)--(iii) limit, and
from the gate of a shared expert, which no protection limits.

\begin{lemma}[What acceptance changes]\label{lem:preserve}
Let $\Delta'$ be a tile after one acceptance step, Equation~\eqref{eq:accept} or~\eqref{eq:protacc}.
(a) $\Delta'(I-\PiC)=\Delta(I-\PiC)$: directions outside the common subspace are not changed.
(b) Under Equation~\eqref{eq:protacc}, also $\Delta'\PiCT=\Delta\PiCT$ for any $M$ and any $W_\lambda$;
the unprotected update changes this component by $-(\Delta-M)W_\lambda\PiCT$.
\end{lemma}
Part (a) keeps the corresponding client-specific parameter components, whereas FedAvg scales a direction
used only by client $k$ by $p_k$ in every round. Part (b) holds for one step and
the subspaces of that round, so whether earlier tasks are preserved over a whole
task is tested empirically. Appendix~\ref{app:proofs} gives a layer-level bound on
forgetting (Theorem~\ref{thm:forgetting}).

\section{Experimental Setup}\label{sec:exp}
\textbf{Data and backbones.} We evaluate on TRACE~\citep{wang2023trace},
training its eight tasks in the order C-STANCE~\citep{zhao2023cstance},
FOMC~\citep{shah2023fomc}, MeetingBank~\citep{hu2023meetingbank}, Py150,
ScienceQA~\citep{lu2022scienceqa}, NumGLUE-cm~\citep{mishra2022numglue},
NumGLUE-ds and 20Minuten~\citep{rios2021minuten}. Each task has 5,000 training
samples in the main setting and 500 in a low-data setting, and we test on the
first 500 test samples of each task with greedy decoding. Qwen3-4B~\citep{yang2025qwen3}
is the main backbone and Llama-3.2-3B-Instruct~\citep{grattafiori2024llama3} the
second.

\textbf{Federated protocol.} Five clients take part in every round. In the
\emph{equal split}, the training samples of a task are dealt to the clients in
turn, so each client holds 1,000 (main) or 100 (low-data) samples. Each task is trained for three communication rounds, with
three local epochs per round, and after each merge a client calibrates its router for 50 steps.

\textbf{Baselines.} \emph{Local} and \emph{FedAvg} use the same client-side learner as \method{},
including expert splitting and gate-row
freezing, and differ only in server aggregation. Local performs
no aggregation, whereas FedAvg replaces each tile held by at
least two clients with its sample-weighted mean.
\emph{Centralized} trains the same learner on the pooled data. \method{}+R is a
variant of our server that also averages the part of each update outside the
agreed subspace and uses $\theta{=}0.5$.

\textbf{Metrics.} We report LA, OP and BWT (Equation~\eqref{eq:metrics}).
Task performance is
measured using accuracy, except for Py150, MeetingBank, and
20Minuten, which use edit similarity, ROUGE-L, and SARI,
respectively. We report means and standard deviations across
the five client models. Because all methods share shards and test
sets, we compare them client by client and claim a difference only if it exceeds
the client spread and at least four of the five clients agree on its sign. Where
fewer than five clients are evaluated, we average those clients and mark the
cell.

\textbf{Ablations and sweeps.} Ablations and sweeps use the low-data setting.
Each ablation trains its own stream with the listed protections switched off, and
hyperparameter sweeps use a four-task stream of C-STANCE, FOMC, MeetingBank and
Py150.\\
\textbf{Code availability.} The implementation of \method{} is available in
APPFL\footnote{\url{https://github.com/APPFL/APPFL}}~\citep{ryu2022appfl,li2025advances},
an open-source framework for federated learning.

\section{Results and Analysis}\label{sec:results}
\subsection{Main Results}\label{sec:overall}
\begin{table*}[htbp]
\caption{Main results on TRACE with Qwen3-4B and 5,000 training samples per task
(1,000 per client, equal split): score on each task right after it is learned
($R_{j,j}$, client mean), LA, OP and BWT. Bold: the higher of the two methods.}
\label{tab:main}
\centering
\footnotesize
\setlength{\tabcolsep}{4.5pt}
\begin{tabular}{@{}l*{8}{c}ccc@{}}
\toprule
Method & C-ST & FOMC & MB & Py150 & SciQA & NG-cm & NG-ds & 20Min & LA & OP & BWT \\
\midrule
FedAvg over tiles & 60.48 & \textbf{65.36} & 37.34 & 56.23 & 79.62 & \textbf{56.92} & 50.77 & \textbf{38.61} & \textbf{55.67} & 49.27 & $-$6.99 \\
\method{} & \textbf{60.68} & 64.84 & \textbf{37.95} & \textbf{56.32} & \textbf{80.47} & 55.13 & \textbf{51.02} & 38.56 & 55.62 & \textbf{51.52} & \textbf{$-$4.68} \\
\bottomrule
\end{tabular}
\vspace{6pt}
\caption{Results on TRACE with Llama-3.2-3B in the setting of Table~\ref{tab:main}:
score on each task right after it is learned ($R_{j,j}$, client mean) and LA.
Bold: the higher of the two methods.}
\label{tab:main_llama}
\begin{tabular}{@{}l*{8}{c}c@{}}
\toprule
Method & C-ST & FOMC & MB & Py150 & SciQA & NG-cm & NG-ds & 20Min & LA \\
\midrule
FedAvg over tiles & 48.80 & \textbf{66.49} & \textbf{41.55} & \textbf{57.79} & \textbf{75.36} & \textbf{53.33} & 48.37 & \textbf{39.81} & \textbf{53.94} \\
\method{} & \textbf{48.88} & 65.04 & 41.28 & 57.24 & 75.16 & 52.56 & \textbf{49.11} & 39.61 & 53.61 \\
\bottomrule
\end{tabular}

\end{table*}

Tables~\ref{tab:main} and~\ref{tab:main_llama} report results with 1,000
samples per client. \method{} matches FedAvg in LA on both backbones (within
0.33 points), and no task differs by more than 1.8 points. Table~\ref{tab:main} reports client means only; the client-spread criterion of \S\ref{sec:exp} has not yet been applied to these runs.

\begin{table}[htbp]
\caption{\method{} minus FedAvg on the accuracy matrix of the runs in
Table~\ref{tab:main} (Qwen3-4B, 5,000 samples per task): score on each task
(column) after learning each task (row).}
\label{tab:griddiff}
\centering\scriptsize
\setlength{\tabcolsep}{2.2pt}
\begin{adjustbox}{max width=\columnwidth}
\begin{tabular}{@{}l*{8}{r}@{}}
\toprule
After & C-ST & FOMC & MB & Py150 & SciQA & NG-cm & NG-ds & 20Min \\
\midrule
T1 & \textbf{$+$0.20} \\
T2 & $+$0.04 & \textbf{$-$0.52} \\
T3 & $+$1.00 & $-$1.66 & \textbf{$+$0.61} \\
T4 & $+$1.25 & $+$0.75 & $+$0.44 & \textbf{$+$0.09} \\
T5 & $+$1.40 & $+$2.82 & $-$0.81 & $+$4.62 & \textbf{$+$0.85} \\
T6 & $+$0.05 & $+$0.26 & $-$1.52 & $+$1.08 & $+$1.10 & \textbf{$-$1.79} \\
T7 & 0.00 & $+$1.61 & $-$0.75 & $+$0.78 & $+$2.23 & $+$5.16 & \textbf{$+$0.25} \\
T8 & $+$1.74 & $+$2.62 & $+$0.08 & $+$3.83 & $+$5.47 & $+$7.05 & $-$0.93 & \textbf{$-$0.05} \\
\bottomrule
\end{tabular}
\end{adjustbox}

\end{table}

Table~\ref{tab:griddiff} compares the full accuracy matrices on Qwen. \method{}
is higher in 27 of the 36 cells, including 22 of the 28 that score an earlier
task after a later one, and the remaining cells differ by at most 1.8 points.
\method{} thus retains more while learning each new task as well as FedAvg.

\subsection{Fewer Samples per Client}\label{sec:lowdata}

\begin{table}[htbp]
\caption{Results with 500 training samples per task, mean $\pm$ standard
deviation over the five clients' models. Centralized is a single model and is
not bolded. Per-task and per-client scores are in Appendix~\ref{app:results}.}
\label{tab:main500}
\centering\footnotesize
\setlength{\tabcolsep}{2pt}
\begin{adjustbox}{max width=\columnwidth}
\begin{tabular}{@{}lccc@{}}
\toprule
Method & LA & OP & BWT \\
\midrule
\multicolumn{4}{@{}l}{\textit{Qwen3-4B, equal split}} \\
\midrule
Local (no merge)        & 50.59\std{0.55} & 44.81\std{2.12} & \textbf{$-$6.61}\std{2.41} \\
FedAvg over tiles       & \textbf{51.38}\std{0.63} & \textbf{45.54}\std{0.89} & $-$6.68\std{1.39} \\
\method{}               & 50.54\std{1.03} & 44.69\std{1.62} & $-$6.67\std{2.35} \\
\midrule
Centralized             & 57.75 & 48.56 & $-$10.50 \\
\midrule
\multicolumn{4}{@{}l}{\textit{Llama-3.2-3B-Instruct, equal split}} \\
\midrule
Local (no merge)        & 51.07\std{1.23} & 49.16\std{0.72} & \textbf{$-$2.19}\std{0.78} \\
FedAvg over tiles       & \textbf{52.78}\std{0.47} & \textbf{50.33}\std{0.27} & $-$2.80\std{0.56} \\
\method{}               & 51.35\std{0.92} & 48.91\std{1.10} & $-$2.79\std{0.86} \\
\method{}+R             & 52.19\std{0.80} & 49.26\std{0.72} & $-$3.35\std{0.92} \\
\midrule
Centralized             & 55.87 & 50.58 & $-$6.04 \\
\bottomrule
\end{tabular}
\end{adjustbox}

\end{table}
With 500 samples per task (100 per client; Table~\ref{tab:main500}), \method{} is 0.8--0.9 (Qwen) and 1.4 (Llama) points below FedAvg in LA and OP with nearly identical BWT. NumGLUE-cm accounts for most of this difference: without it, the two methods perform comparably on Qwen and within 0.6 points on Llama (Table~\ref{tab:nong}).
On this task, clientson Llama-3.2-3B agree on only 22--38\% of their updates after five earlier
tasks, so \method{} merges conservatively (Appendix~\ref{app:server}); with
1,000 samples per client, the difference falls to 1.8 points or less.
\method{}+R, which also averages the update outside the agreed subspace, achieves higher LA than \method{} on all five clients and matches FedAvg's LA without
NumGLUE-cm.

\subsection{Ablations}\label{sec:ablation}
\begin{table}[htbp]
\caption{Protection ablations on Llama-3.2-3B (500 samples per task, equal
split). Each row trains its own stream; all rows use clients 0--3, the clients
evaluated in every run. Bold: best configuration of our server. Merge protection
alone, evaluated on two clients, is in Appendix~\ref{app:ablation}.}
\label{tab:ablation}
\centering\footnotesize
\setlength{\tabcolsep}{2pt}
\begin{adjustbox}{max width=\columnwidth}
\begin{tabular}{@{}lccc@{}}
\toprule
Configuration & LA & OP & BWT \\
\midrule
\method{} (both protections) & \textbf{51.38}\std{1.06} & 49.16\std{1.09} & $-$2.53\std{0.75} \\
\quad gate freezing only     & 50.48\std{1.28} & \textbf{49.41}\std{0.50} & \textbf{$-$1.22}\std{1.46} \\
\quad no protection          & 50.42\std{1.14} & 48.74\std{0.85} & $-$1.92\std{1.40} \\
\midrule
Local (no merge)             & 51.09\std{1.42} & 49.30\std{0.75} & $-$2.05\std{0.81} \\
FedAvg over tiles            & 52.83\std{0.53} & 50.38\std{0.28} & $-$2.80\std{0.65} \\
\bottomrule
\end{tabular}
\end{adjustbox}
\end{table}

\textbf{Protections.} Table~\ref{tab:ablation} disables the protections on
Llama. Gate freezing is protection (i), and merge protection is (ii) and (iii) together. Merge protection mainly helps learning: without it, MeetingBank and
Py150 are learned 2.6--2.9 points worse. Gate freezing alone gives the best BWT on clients 0--3 ($-1.22$, against $-2.53$ with both protections and $-1.92$ with none), largely because Py150 keeps its score instead of losing 8.2 points as it does with both protections (Table~\ref{tab:abl_task}); dropping merge protection costs 0.9 points of LA. These BWT differences lie within the client spread, gate freezing leaves BWT unchanged on clients 0--1, and adding merge protection lowers BWT in every client subset we evaluated (Table~\ref{tab:abl_more}), so in these runs merge protection improves learning but does not reduce forgetting. FOMC, MeetingBank and ScienceQA change by
similar amounts under every configuration, including local training, so these
changes come from the client side rather than the merge. \\
\textbf{Settings and scale.} On the four-task stream (Qwen), \method{} is 0.4 points above
FedAvg in LA, and the merge settings (agreement threshold, conflict threshold, change LA by at most 0.3 points (Table~\ref{tab:sweep}). The
budget matters more: increasing the tile budget from 15\% to 25\% improves LA by 3.3 points, whereas reducing it to 12\% lowers LA by 0.5 points. An absolute size floor and writing the server labels back into the
expert structure add 1.3 and 1.7 points, but we have measured them only on LA.
With ten clients of 50 samples each, both aggregation methods exhibit substantial performance degradation on MeetingBank and Py150 (MeetingBank by about 12.5 points and Py150 by 21.9 (\method{}) and 20.7 (FedAvg) points, relative to five clients) and end within 0.7 points of each other; a similar decline occurs with one communication round per task, suggesting that limited local training contributes to the degradation. Measuring client updates from the start of the task instead of the
start of the round changes no metric by more than 0.2 points
(Table~\ref{tab:abl_ref}).
\section{Conclusion}\label{sec:conclusion}
\method{} supports federated continual learning through a mixture of
sparse experts on a frozen language model. Task-aware tile selection
and selective expert reuse organize knowledge across successive tasks,
while \SEAM{} aggregates expert content across clients along agreed update directions and retains client-specific components outside them. Three protections target routing drift and merge erasure, including exact preservation of tile components in a selected prior-overlap subspace during each merge; in our ablation, merge protection mainly improved learning rather than retention. Results on TRACE show comparable learning accuracy to FedAvg in the main setting, with available evaluations indicating better retention. With fewer samples per client, FedAvg's broader averaging gives slightly higher
learning accuracy.
\section*{Limitations}
Clients share model updates, which can leak information about their data, so combining \method{} with
secure aggregation or differential privacy~\citep{abadi2016dpsgd} remains future work. We would also
like to explore domain adaptation in federated continual learning, where the clients' data shift from
one domain to another over time. Our clients hold near-identically distributed shards, and evaluating
\method{} with non-IID clients is left to future work.
\section*{Acknowledgment}
This work was supported by the U.S. Department of Energy, Office of Science, under contract number DE-AC02-06CH11357. This research also used resources of the Argonne Leadership Computing Facility, which is a U.S. Department of Energy Office of Science User Facility operated under contract DE-AC02-06CH11357.
\bibliography{custom}
\section*{Appendix}

\section{Method Details}\label{app:method}
This appendix gives the details of each step; Appendix~\ref{app:hyper} lists all settings.

\begin{algorithm}[htbp]
\caption{\method{}: one task}
\label{alg:fedseam}
\small
\begin{algorithmic}[1]
\State clients: warm-up on $D_t^k$; column correction, novelty, budget; own tile selection
\State clients: direction-aware split; snapshots $\hat W$
\For{$r = 1$ \textbf{to} $R$}
  \State clients: train Eq.~\eqref{eq:loss}, earlier gate rows frozen \hfill (i)
  \State clients: send the tiles of the current task's experts
  \State server: bin tiles by coordinate; single holders stay local; guard; weights $\pi_k$
  \State per bin: $\QC$ from the spectrum; $\mathcal{T}$ from earlier experts and snapshots \hfill (ii)
  \State merged $M$, reduced along $\mathcal{C}\cap\mathcal{T}$; signed test
  \State clients: protected acceptance, Eq.~\eqref{eq:protacc} \hfill (iii)
  \State clients: calibrate gate rows of the current task
\EndFor
\State evaluate $R_{t,j}$ on every client's final model
\end{algorithmic}
\end{algorithm}

\textbf{Keeping the answer.} A sample is a prompt $p$ ending in a response cue, followed by an answer $a$.
If $|p|+|a|\le n_{\max}$ it is left unchanged. Otherwise the answer is kept and the prompt keeps its first
$h=n_{\max}-|a|-s$ and last $s=48$ tokens, which contain the instruction and the cue. Only if the answer
leaves fewer than $64$ prompt tokens is the answer shortened.

\textbf{Warm-up.} After step $s$ we compute
\[
d_s=\norm{G^{(s)}-G^{(s-1)}}\big/\norm{G^{(s)}}
\]
and stop once $d_s\le\mu_s+\sigma_s$ over the five preceding values has held for ten steps (fifteen
for generation tasks). With two instances per step, the warm-up runs between half a pass and one pass over
the shard ($25$ to $50$ steps for a shard of $100$ instances in the low-data setting, $250$ to $500$ for the $1{,}000$ of the main setting), so it sees each instance at most once. The client keeps the gradients
of its selected tiles, which the split uses as evidence.

\textbf{Budget walk.} Layer $\ell$ may receive
\[
\max\Big\{1,\Big\lfloor\kappa B\,v_\ell\Big/\textstyle\sum_{\ell'}v_{\ell'}\Big\rfloor\Big\},\quad
v_\ell=\frac{1}{\mathrm{cov}_\ell+0.1}
\]
tiles, where $\mathrm{cov}_\ell$ is the used fraction of the layer. The walk returns only the number $N_t$; the
selected tiles are the $N_t$ highest scores. With $\kappa{=}1$, as in the main configuration, the walk
takes exactly $B$ tiles; with $\kappa{=}1.5$, as in the structure-only pass of
Appendix~\ref{app:structure}, the first task, which has no used tiles, stops at the third record point
after $B$. Layers with fewer than $\lceil T\rceil$ selected tiles are dropped. Usage counts are recomputed from
stored selections.

\textbf{Direction test.} $\mu_0(x)$ and $\sigma_0(x)$ are computed from up to 32
tiles that the new task selected in the same layer, or from random subspaces of
the same ranks when fewer than four such tiles exist. If expert $e$ has not
changed the tile, the earlier task's own warm-up gradient replaces
$-\Delta W_e^{x}$, and a tile for which no score can be computed is shared
without testing.

\textbf{Split.} An untrained expert, which occurs only in structure-only runs, uses its warm-up gradient
as evidence. A tile without evidence is shared. When one coordinate is held by several earlier experts
and passes for each of them, the shared expert holds it once, with the values of the gate origin (zero if
the origin does not hold it), and the other experts keep their own values (Algorithm~\ref{alg:split}).
The shared expert starts from the gate row of the gate origin.

\textbf{Training.} The anchor $\hat W$ is taken once per task and kept for every round; retaking it each
round would let the parameters drift in small steps without penalty. Calibration trains only the gate rows of the current task for $50$
steps.

\textbf{Agreement spectrum.} Before the spectrum is computed, tiles whose norm
exceeds three times the median norm in the bin are scaled down to that bound,
and tiles that are not finite are dropped. The eigenvalues $\lambda_j$ are
multiplied by the fraction of the bin's clients whose update is nonzero, so a
client that did not move counts as absent. Over the rounds of a task, the sorted
spectrum is averaged with the smoothed spectrum of the previous round, with
weight $0.5$; the weights recomputed after the sign test are not smoothed. For
fixed $\pi_k$, $\lambda_j$ depends only on the client subspaces and does not
change when an update is rescaled; $\pi_k$ itself does, because rescaling one
update changes $\bar U$.

\textbf{Bin labels and structural variant.} For analysis, each bin receives a
label: \emph{client-private} if the client tiles keep on average less than half
of their energy inside $\mathcal{C}$; otherwise \emph{task-shared} if $M$, after
the reduction in Equation~\eqref{eq:merged}, keeps at least $10\%$ of its energy
in $\mathcal{C}\cap\mathcal{T}$, and \emph{task-global} if not. An optional
variant writes the labels back as structure by moving the residual
$\Delta^k(I-\PiC)$ into a private expert; this decomposition is exact
(Proposition~\ref{prop:emit}), and the variant serves only as an ablation. The
signed test of \S\ref{sec:protect} changes client tiles only through the unequal
weights in $W_\lambda$, because (iii) already excludes
$\mathcal{C}\cap\mathcal{T}$ from the pull.

\begin{algorithm*}[htbp]
\caption{Budget walk of the task-aware selection (sets the number of tiles)}
\label{alg:walk}
\small
\begin{algorithmic}[1]
\Require scores $s$; usage counts $u$ and used set $\mathcal{U}$ from earlier tasks; base budget $B$; expansion limit $\kappa$; task index $t$
\State compute the layer limits;\; $\tau\gets\max\{0.005,0.05/t\}$;\; $\mathcal{A}\gets\emptyset$;\; $n\gets 0$;\; $n_{\mathrm{new}}\gets 0$;\; $H\gets[\,]$
\For{each tile $x$ in order of decreasing $s(x)$}
  \If{the layer of $x$ is full \textbf{or} $u(x)\ge 2$} \textbf{continue} \EndIf
  \State $\mathcal{A}\gets\mathcal{A}\cup\{x\}$;\; $n\gets n+1$;\; \textbf{if} $x\notin\mathcal{U}$ \textbf{then} $n_{\mathrm{new}}\gets n_{\mathrm{new}}+1$
  \If{$n\ge B$ \textbf{and} $n \bmod 50 = 0$}
     append $n_{\mathrm{new}}/n$ to $H$;\; \textbf{if} the last three entries of $H$ differ by less than $\tau$ \textbf{then break}
  \EndIf
  \If{$n\ge \kappa B$} \textbf{break} \EndIf
\EndFor
\State \Return $N_t\gets n$ \Comment{the task takes the $N_t$ highest-scoring tiles}
\end{algorithmic}
\end{algorithm*}

\begin{algorithm}[htbp]
\caption{Split of one client at the start of task $t$}\label{alg:split}
\begin{algorithmic}[1]
\Require tiles $\mathcal{S}_t^\ell$ selected by task $t$ in each layer $\ell$ (layers with fewer than $T$
  selected tiles removed); earlier experts $e$ with tiles $\mathcal{X}_e^\ell$, values $\Delta W_e$ and gate
  rows $(g_e,b_e)$
\State $\mathcal{N}^\ell\gets\mathcal{S}_t^\ell$ for every layer $\ell$
  \Comment{tiles of the new task's own expert}
\For{each earlier expert $e$}
  \For{each layer $\ell$}
    \State $\mathcal{P}\gets\{x\in\mathcal{X}_e^\ell\cap\mathcal{S}_t^\ell:$ $x$ passes
      Equations~\eqref{eq:null} and~\eqref{eq:sign}, or no score exists for $x\}$
    \If{$|\mathcal{P}|\ge T$ \textbf{and} $|\mathcal{X}_e^\ell\setminus\mathcal{P}|\in\{0\}\cup[T,\infty)$
        \textbf{and} $|\mathcal{N}^\ell\setminus\mathcal{P}|\in\{0\}\cup[T,\infty)$}
      \State $\mathcal{I}_e^\ell\gets\mathcal{P}$ \Comment{tiles $e$ passes to the shared expert}
    \Else
      \State $\mathcal{I}_e^\ell\gets\emptyset$ \Comment{$e$ keeps all; the new task keeps its own copies}
    \EndIf
  \EndFor
  \State $\mathcal{N}^\ell\gets\mathcal{N}^\ell\setminus\mathcal{I}_e^\ell$ for every layer $\ell$
\EndFor
\For{each layer $\ell$ with $\mathcal{I}_e^\ell\neq\emptyset$ for some $e$}
  \State $o\gets\arg\max_e|\mathcal{I}_e^\ell|$, the first such expert on ties \Comment{gate origin}
  \State shared expert $s$ in layer $\ell$: tiles $\bigcup_e\mathcal{I}_e^\ell$, gate row $(g_o,b_o)$,
    values $\Delta W_o$ on $\mathcal{I}_o^\ell$ and $0$ on all other tiles
  \State $\mathcal{X}_o^\ell\gets\mathcal{X}_o^\ell\setminus\mathcal{I}_o^\ell$; every other earlier
    expert keeps $\mathcal{X}_e^\ell$ and its values
\EndFor
\State new expert of task $t$: tiles $\mathcal{N}^\ell$, values $0$, a new gate row
\State every earlier expert keeps its gate row; an expert with no tiles left in a layer leaves that layer
\end{algorithmic}
\end{algorithm}

\section{Measurements of Position-Based Sharing}\label{app:prelim}
These measurements come from a federated version of the SETA
learner~\citep{siddika2026seta} on Qwen3-4B with five clients, which selects
tiles by raw gradient magnitude and shares them by position. They motivate the
column correction, the size threshold and the truncation rule (Appendix~\ref{app:method}).

\textbf{Selection by raw gradient magnitude.} The gradient of a weight is the
product of an output error and an input activation. A few hidden channels carry
large activations for every input, and the tiles that read them form whole
columns of the $8\times20$ grid. With $24$ tiles per layer, a selection by raw
rank tends to choose three columns of eight tiles for every task. At the second
task, $763$ of the $864$ selected tiles ($88\%$) had been selected by the first
task. Two independent selections of $24$ of the $160$ tiles in a layer would
overlap by $24/160=15\%$ in expectation.

\textbf{Absolute size floors.} Its split required at least five
tiles per layer and at least $15\%$ of the budget ($\approx130$ tiles) for the
whole expert. On one client, $324$ tiles were accepted for sharing, $38$
survived the per-layer floor, and $38$ is below the expert minimum, so the split
was refused. Every task on every client ended without a split. The single
threshold $T=\phi B$ of \S\ref{sec:split} replaces both floors.

\textbf{Truncation.} With a window of $2{,}048$ tokens, $46$ of $100$
MeetingBank samples of one client reached the limit, and $44$ held no answer
token after truncation. The current rule keeps the whole answer and removes the
middle of the prompt.

\section{Further Theory and Proofs}\label{app:proofs}
\begin{theorem}[Forgetting at a layer]\label{thm:forgetting}
Consider the model of one client, one layer, and a task $j$. Let $x$ be the input of this layer for a
sample of task $j$, held fixed. For a later task $m$, let $\mathcal{E}_m$ be the experts that the split of
task $m$ creates in this layer (the shared expert and the task's own expert), $\hat W_e$ the snapshot of
$e\in\mathcal{E}_m$ taken after the split, $\Delta W_e^{(m)}$ its value at the end of task $m$, and
$\sigma_e^{(m,0)}$ and $\sigma_e^{(m)}$ its gate at the start and at the end of task $m$. Write
$\omega_e=\Delta W_e^{(m)}-\hat W_e$ for its weight drift and
$\rho_e(x)=\sigma_e^{(m)}(x)-\sigma_e^{(m,0)}(x)$ for its gate change. With sigmoid gating,
protection (i) without weight decay on the gate rows, and the output-preserving split
(Proposition~\ref{prop:splitpres}), for every $t>j$,
\begin{align}
y^{(t)}(x)-y^{(j)}(x)
&=\sum_{m=j+1}^{t}\sum_{e\in\mathcal{E}_m}\Big[\sigma_e^{(m)}(x)\,\omega_e\,x\nonumber\\
&\qquad\quad+\rho_e(x)\,\hat W_e\,x\Big],
\label{eq:forgetting}
\end{align}
and hence
\begin{align}
\norm{y^{(t)}(x)-y^{(j)}(x)}
&\le\sum_{m=j+1}^{t}\sum_{e\in\mathcal{E}_m}\Big[\norm{\omega_e\,x}\nonumber\\
&\qquad\quad+|\rho_e(x)|\,\norm{\hat W_e\,x}\Big].
\label{eq:forgettingbound}
\end{align}
If the loss $F_j$ of task $j$ on this sample is $L$-Lipschitz in the output of this layer, with all
other layers held fixed, then $|F_j^{(t)}-F_j^{(j)}|$ is at most $L$ times the right-hand side of
Equation~\eqref{eq:forgettingbound}.
\end{theorem}

\begin{proof}
At the fixed input $x$, write $y^{(t)}-y^{(j)}=\sum_{m=j+1}^{t}\big(y^{(m)}-y^{(m-1)}\big)$, where
$y^{(m)}$ is the output at the end of task $m$. The split at the start of task $m$ leaves the output
unchanged, $y^{(m,0)}=y^{(m-1)}$ (Proposition~\ref{prop:splitpres}). During task $m$, local training,
acceptance and calibration change only the experts of task $m$, that is $\mathcal{E}_m$, and their gate
rows: the experts of earlier tasks are frozen, and by (i) their gate rows receive no gradient, so without
weight decay these rows, and with sigmoid gating the corresponding gates at $x$, do not change. The anchoring penalty acts on expert values, not on gate rows, so it does not affect this step. Since the
snapshot is taken after the split, $\Delta W_e^{(m,0)}=\hat W_e$, and therefore
\begin{align*}
y^{(m)}-y^{(m,0)}
&=\sum_{e\in\mathcal{E}_m}\Big[\sigma_e^{(m)}(x)\,\Delta W_e^{(m)}\\
&\qquad\quad-\sigma_e^{(m,0)}(x)\,\hat W_e\Big]x .
\end{align*}
Adding and subtracting $\sigma_e^{(m)}(x)\,\hat W_e x$ and summing over $m$ gives
Equation~\eqref{eq:forgetting}. The bound follows from the triangle inequality and
$0<\sigma_e^{(m)}(x)<1$, and the statement about $F_j$ from the Lipschitz assumption.
\end{proof}

\paragraph{What controls each term.}
For the task's own expert, $\hat W_e=0$, so its first term is its whole output: the intrusion of a new
expert. For a shared expert, the first term is the drift of its inherited and new tiles from the
snapshot. Within task $m$ this drift comes from local training and from acceptance. By
Lemma~\ref{lem:preserve}(b), each acceptance step leaves the component of a tile in that round's
$\mathcal{C}\cap\mathcal{T}$ unchanged, so acceptance adds drift only along directions of $\mathcal{C}$
whose squared cosine with $\mathcal{T}$ is below $0.5$. The penalty in Equation~\eqref{eq:loss} limits
local training: at a stationary point of the local objective,
$\fro{\Delta W_e-\hat W_e}=\fro{\nabla_{\Delta W_e}\mathcal{L}_{\mathrm{task}}}/(2\mu)$. Local training stops after a
few epochs and acceptance moves the weights afterwards, so this indicates the scale of the drift rather
than bounding it in our runs. The second term is the change in routing of the values a shared expert
inherited; it is zero at the split and grows as the shared expert's gate row trains, and no protection
limits it. The theorem holds for a fixed layer input; across layers, the input $x$ changes as well.

\begin{proposition}[Column invariance]\label{prop:colinv}
Let $m'(\ell,r,q)=a_q\,m(\ell,r,q)+c_q$ with $a_q>0$. As $\varepsilon\to0$, $\tilde m$ of
Equation~\eqref{eq:fixcol} is the same for $m$ and $m'$.
\end{proposition}
\begin{proof}
The column mean and deviation of $m'$ are $a_q\mu_{\ell,q}+c_q$ and $a_q\sigma_{\ell,q}$, so the standardised
values coincide.
\end{proof}

\begin{proposition}[The split preserves the output]\label{prop:splitpres}
Let a layer compute $y(x)=W_0x+\sum_e\sigma(g_e^{\top}x+b_e)\,\Delta W_e\,x$ with sigmoid gates, where
$\Delta W_e$ is zero outside the tiles of expert $e$. After Algorithm~\ref{alg:split}, $y(x)$ is
unchanged for every input $x$, for any number of earlier experts that pass tiles to the shared expert.
\end{proposition}
\begin{proof}
Fix a layer and its gate origin $o$. For each earlier expert $e$, let $\mathcal{R}_e$ be the tiles it
keeps and $\mathcal{I}'_e$ the tiles whose values it passes on, so that
$\Delta W_e=\Delta W_e|_{\mathcal{R}_e}+\Delta W_e|_{\mathcal{I}'_e}$. By construction
$\mathcal{I}'_o=\mathcal{I}_o^\ell$ and $\mathcal{I}'_e=\emptyset$ for every $e\neq o$: the other experts
keep all their tiles, including those that also enter the shared expert. After the split, each earlier
expert contributes $\sigma(g_e^{\top}x+b_e)\,\Delta W_e|_{\mathcal{R}_e}x$ with its gate row unchanged;
the shared expert contributes $\sigma(g_o^{\top}x+b_o)\,\Delta W_o|_{\mathcal{I}_o^\ell}x$, because its
values are zero on all other tiles and its gate row equals that of $o$; and the new expert contributes
nothing, because its values are zero. The sum of these contributions is
$\sum_e\sigma(g_e^{\top}x+b_e)\,\Delta W_e\,x$. With sigmoid gates, each gate depends only on its own row,
so the added rows do not change the existing gates, and $y(x)$ is unchanged.
\end{proof}
The proposition relies on the gates being independent: under softmax gating, adding gate rows would
rescale the existing gates. It also shows why the gate origin is needed: if the shared expert inherited
the values of several earlier experts $e$ with different gate rows, their tiles would be gated by
$\sigma(g_o^{\top}x+b_o)$ instead of $\sigma(g_e^{\top}x+b_e)$ after the split, and the output would
change.

\begin{proposition}[Emission is an exact decomposition]\label{prop:emit}
In the optional emission variant, $\Delta^k=\Delta^k\PiC+\Delta^k(I-\PiC)$ and the two parts act on orthogonal input subspaces. If the
residual is kept and new gate rows equal the row of the parent, the output is unchanged. A residual
dropped by the floor $\nu$ changes the tile by at most $\sqrt{\nu}\fro{\Delta^k}$.
\end{proposition}
\begin{proof}
$\PiC$ and $I-\PiC$ are complementary orthogonal projectors. A residual is dropped only if
$\fro{\Delta^k(I-\PiC)}^2<\nu\fro{\Delta^k}^2$.
\end{proof}

\textbf{Discussion of Proposition~\ref{prop:decomp}.} The identity holds at one
layer for a fixed layer input. In the full network, earlier layers can change
$x$, and hence the gate values, even when this layer's gate rows are frozen. For
a fixed layer input, protection (i) removes the first term: every expert present
at the end of task $j$ belongs to a task no later than $j$, so its gate row
receives no gradient afterwards, and a split keeps the gate row of an expert's
remaining tiles. A later split that moves tiles of such an expert into a new
shared expert changes the second and third terms by equal and opposite amounts,
so the split itself does not change $\Delta y$. After the split, those tiles are
trained under the shared expert's weights and gate row. Their contribution to
$\Delta y$ is then the change of their weights, which the penalty in
Equation~\eqref{eq:loss} and protections (ii) and (iii) limit, together with the
difference between the shared expert's gate and the original one, which no
protection limits.

\textbf{Proof of Proposition~\ref{prop:decomp}.} Write
$y^{(s)}=W_0x+\sum_{e\in A_s}\sigma_e^{(s)}(x)\Delta W_e^{(s)}x$ with $A_s$ the experts present at $s$. Subtract the
two checkpoints, add and subtract $\sum_{e\in A_j}\sigma_e^{(t')}(x)\Delta W_e^{(j)}x$ and group. Frozen unique
experts contribute zero to the weight term. \qed

\textbf{Discussion of Lemma~\ref{lem:preserve}.} The guarantee concerns a single
acceptance step and is exact for the subspaces of that round. Across rounds,
$\mathcal{C}$ and $\mathcal{T}$ are re-estimated, and local training between
rounds can change the protected component, so the lemma does not by itself bound
the change over a task; Theorem~\ref{thm:forgetting} accounts for both sources of
drift. Part (a) means acceptance never alters directions outside the common
subspace, so what a client alone learned is kept; this is the mechanism behind RQ1, which we assess through task scores rather than measure directly. Part (b) means a single acceptance step does not alter the component of
a shared tile in $\mathcal{C}\cap\mathcal{T}$; this is the mechanism behind RQ2, which the protection ablation (Table~\ref{tab:ablation}) tests only indirectly.

\textbf{Proof of Lemma~\ref{lem:preserve}.} (a) Because $\QC^{\top}\QC=I$,
\[
W_\lambda(I-\PiC)=\QC\,\mathrm{diag}(\lambda_j)\,\QC^{\top}(I-\QC\QC^{\top})=0.
\] Since $\mathcal{C}\cap\mathcal{T}\subseteq\mathcal{C}$, $(I-\PiCT)(I-\PiC)=I-\PiC$, so the
protected map also vanishes on $I-\PiC$. For FedAvg, if $\Delta^jv=0$ for all $j\ne k$, the mean applied to
$v$ equals $p_k\Delta^kv$. (b) Multiply Equation~\eqref{eq:protacc} by $\PiCT$ from the right.
Since $(I-\PiCT)\PiCT=0$, the second term vanishes for any $M$, $W_\lambda$ and client. For the unprotected map the change is $-(\Delta-M)W_\lambda\PiCT$. When the
reduction has driven $M\PiCT$ close to zero while $\Delta\PiCT\ne0$, the pull moves that content towards
zero at a rate set by the eigenvalues of $W_\lambda$, which is merge erasure. \qed

\section{Settings}\label{app:hyper}
Table~\ref{tab:hyper} lists the settings of the main configuration. All runs are tested on the first
$500$ test instances of every task (all if fewer).

\begin{table*}[htbp]
\caption{Settings of the main configuration.}
\label{tab:hyper}
\centering
\small
\begin{tabular}{@{}l p{0.44\textwidth} p{0.38\textwidth}@{}}
\toprule
Group & Setting & Value \\
\midrule
Federation & clients, rounds per task, local passes per round & 5, 3, 3 \\
           & training instances per client and task & 1{,}000 (100 in the low-data setting) \\
Architecture & adapted matrices; tile side & value projections; 128 \\
             & budget per task; expansion limit $\kappa$ & 15\% (864 tiles); 1 \\
             & router & sigmoid, initial bias 1.0 \\
Warm-up & steps; instances per step & at most one pass; 2 \\
        & learning rate & $9.65\times10^{-6}$; $2\times10^{-6}$ for generation tasks \\
Selection & skip if used by; record interval; minimum tiles per layer & 2 earlier tasks; 50 tiles; $\lceil T\rceil$ \\
Split & energy; rank limit; chance samples; margin & 0.9; 8; 32; 3 standard deviations \\
            & $\phi$; size threshold $T=\phi B$ & 0.015; 12.96 ($\lceil T\rceil=13$) tiles \\
Training & learning rate; warm-up steps; clipping & $9.65\times10^{-6}$; 2; 1.0 \\
         & instances per step; anchor weight $\mu$ & 8; 0.1 \\
         & maximum length & 512; 768 for 20Minuten; 2{,}048 for MeetingBank (1{,}536 for Llama-3.2-3B) \\
\SEAM{} & energy; rank limit; $\theta$; chance margin & 0.9; 32; 0.7; 0.05 \\
        & label thresholds; $c_{\mathrm{conf}}$; smoothing over rounds & 0.5 and 0.1; $-0.1$; 0.5 \\
        & minimum clients per bin; guard factor; private floor $\nu$ (emission variant) & 2; 3; 0.10 \\
Calibration & steps; learning rate & 50; $10^{-4}$ \\
\bottomrule
\end{tabular}

\end{table*}

\section{Efficiency and Server-side Scores}\label{app:efficiency}
\textbf{Parameters.} A task trains at most $\kappa B$ tiles of $128\times128$ values. With $B{=}864$ and
$\kappa{=}1$ this is $14.2$M values, $0.36\%$ of the $3.94$B parameters of Qwen3-4B, plus one gate row per
new expert and layer. Finished experts are kept, so the stored state grows with the number of tasks; in
the structure-only pass of Table~\ref{tab:sos_summary}, all experts together held $2{,}933$ tiles after
eight tasks ($48$M values, $1.2\%$ of the backbone). LoRA of rank $8$ on the same value projections trains
about $1.0$M values.
\textbf{Communication.} After each round a client sends the tiles of the current task's experts, at most
$\kappa B$ tiles. Once per task it also sends their start-of-task values and, for every merged
coordinate, the $128\times128$ Gram matrix of the tiles its earlier experts hold there, which the prior
subspace needs. Neither grows with the index of the task.
\textbf{Computation.} The spectrum costs one singular value decomposition per client and bin and one
eigen decomposition per bin on matrices of size $128$.
\textbf{Server-side scores.} A model assembled on the server carries a router taken from one client
before calibration; in earlier runs its scores were at chance level ($3$ to $7\%$ on the eighth
task) while client models scored about $40\%$, so all reported cells are client averages.

\textbf{Reproducibility.} Partitions, task order and seeds are fixed and scripted. The code base contains
tests that run without a graphics processor and check the selection, the split, the truncation rule and
the metrics. The code is available in the anonymous repository linked in the abstract; configurations and full accuracy matrices will be released with it.

\section{Additional Results}\label{app:results}

\subsection{Router Calibration}\label{sec:router_calib}

\begin{figure*}[t]
\centering
\includegraphics[width=0.8\textwidth]{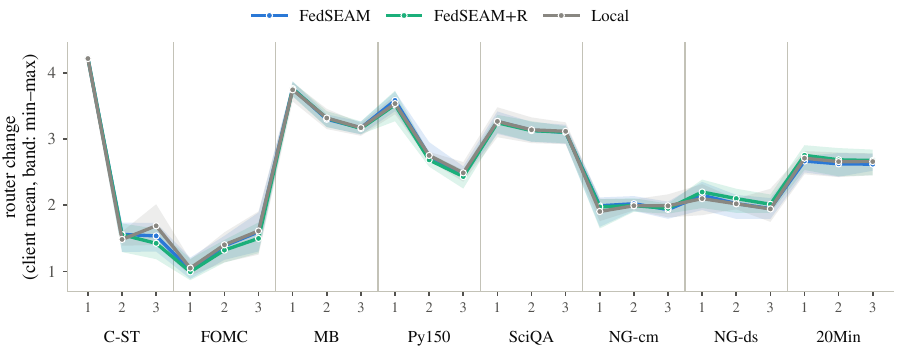}\\[2pt]
{\small (a) Qwen3-4B}\\[6pt]
\includegraphics[width=0.8\textwidth]{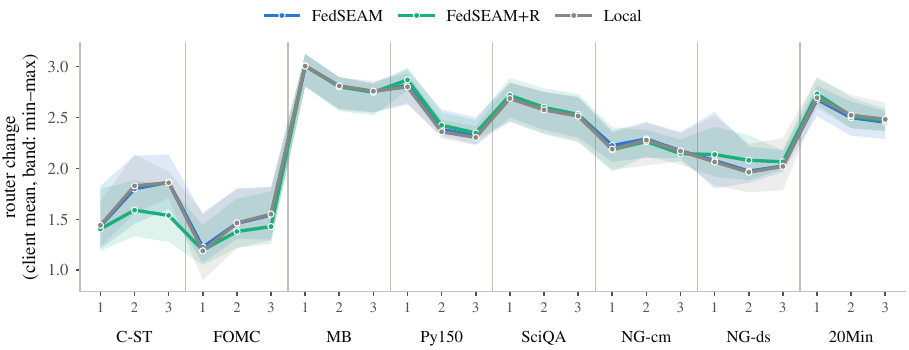}\\[2pt]
{\small (b) Llama-3.2-3B}
\caption{Router change made by calibration in each task and round: the norm of the difference between a
client's router after calibration and after local training. Lines are client means; bands span the
five clients.}
\label{fig:router_change}
\end{figure*}

After the server merge, each client recalibrates its router on its own data.
Figure~\ref{fig:router_change} shows how far calibration moves the router in every task and round. The
size of the step depends on the task: it is largest on MeetingBank and smallest on FOMC, and in most
tasks it shrinks from the first to the third round. The three methods nearly overlap on both models,
including Local, which calibrates without any merge. Calibration therefore re-fits routing to each
task's data at a roughly fixed cost per round, independent of how much the merge changed the experts.

\subsection{Backward Evaluation with 5,000 Samples per Task}\label{app:robust}

\begin{figure}[htbp]
\centering
\includegraphics[width=\columnwidth]{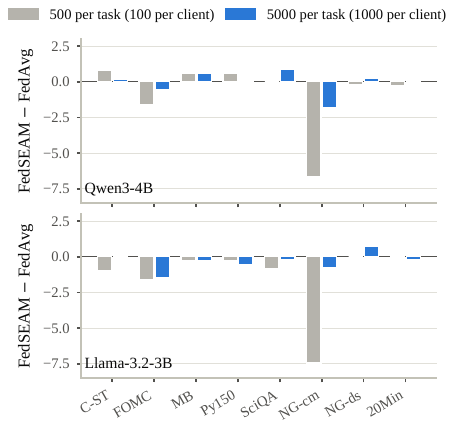}
\caption{\method{} minus FedAvg on each task right after it is learned
($R_{j,j}$), with 500 and with 5,000 training samples per task.}
\label{fig:datasize}
\end{figure}

\begin{figure}[htbp]
\centering
\includegraphics[width=\columnwidth]{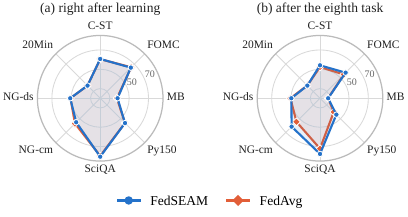}
\caption{Per-task scores of \method{} and FedAvg on Qwen3-4B with 5,000 training
samples per task (client mean). (a) Score right after each task is learned.
(b) Score of the model after the eighth task. The radial axis starts at 20.}
\label{fig:radar5000}
\end{figure}

Figure~\ref{fig:datasize} compares the difference between \method{} and FedAvg on each
task right after it is learned at both data sizes. Figure~\ref{fig:radar5000} shows the per-task scores of both methods with 5,000 samples per task. Tables~\ref{tab:grid_seam}
and~\ref{tab:grid_avg} give the backward evaluation of the Qwen3-4B runs of
Table~\ref{tab:main}; the off-diagonal cells score an earlier task
with the model after a later one.

\begin{table*}[!t]
\caption{Backward evaluation of \method{}, Qwen3-4B, 5,000 samples per task:
score on each task (column) after learning each task (row). Unmarked cells
average all five clients; a superscript gives the number of clients averaged.}
\label{tab:grid_seam}
\centering\small
\setlength{\tabcolsep}{4.5pt}
\begin{tabular}{@{}l*{8}{c}@{}}
\toprule
After $\backslash$ on & C-STANCE & FOMC & MeetingBank & Py150 & ScienceQA & NumGLUE-cm & NumGLUE-ds & 20Minuten \\
\midrule
T1 & 60.68 &  &  &  &  &  &  &  \\
T2 & 59.20 & 64.84 &  &  &  &  &  &  \\
T3 & 54.76 & 59.23 & 37.95 &  &  &  &  &  \\
T4 & 57.50 & 55.34 & 29.81 & 56.32 &  &  &  &  \\
T5 & 49.90 & 52.82 & 28.70 & 33.58 & 80.47 &  &  &  \\
T6 & 54.35 & 55.70 & 31.13 & 53.52 & 82.50 & 55.13 &  &  \\
T7 & 53.80 & 58.97 & 31.57 & 54.89 & 80.83 & 59.65 & 51.02 &  \\
T8 & 54.07 & 57.46 & 28.44 & 43.94 & 77.61 & 61.54 & 49.69 & 38.56 \\
\bottomrule
\end{tabular}

\end{table*}

\begin{table*}[!t]
\caption{Backward evaluation of FedAvg over tiles, Qwen3-4B, 5,000 samples per
task, as in Table~\ref{tab:grid_seam}.}
\label{tab:grid_avg}
\centering\small
\setlength{\tabcolsep}{4.5pt}
\begin{tabular}{@{}l*{8}{c}@{}}
\toprule
After $\backslash$ on & C-STANCE & FOMC & MeetingBank & Py150 & ScienceQA & NumGLUE-cm & NumGLUE-ds & 20Minuten \\
\midrule
T1 & 60.48 &  &  &  &  &  &  &  \\
T2 & 59.16 & 65.36 &  &  &  &  &  &  \\
T3 & 53.76 & 60.89 & 37.34 &  &  &  &  &  \\
T4 & 56.25 & 54.59 & 29.37 & 56.23 &  &  &  &  \\
T5 & 48.50 & 50.00 & 29.51 & 28.96 & 79.62 &  &  &  \\
T6 & 54.30 & 55.44 & 32.65 & 52.44 & 81.40 & 56.92 &  &  \\
T7 & 53.80 & 57.36 & 32.32 & 54.11 & 78.60 & 54.49 & 50.77 &  \\
T8 & 52.33 & 54.84 & 28.36 & 40.11 & 72.14 & 54.49 & 50.62 & 38.61 \\
\bottomrule
\end{tabular}

\end{table*}

\subsection{Per-Task Results}\label{app:pertask_res}
\begin{figure*}[!t]
\centering
\includegraphics[width=\textwidth]{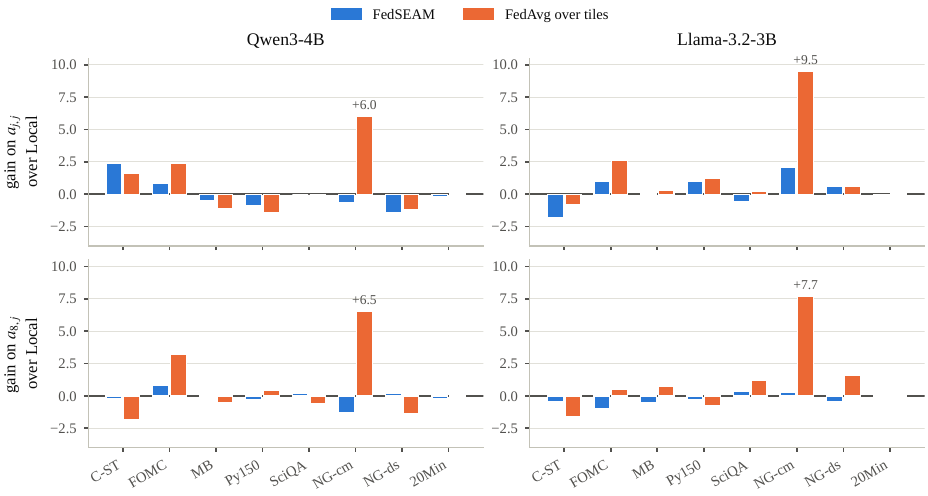}
\caption{Gain over local training on each task (500 samples per task, equal
split, client mean). Top:
right after the task is learned ($R_{j,j}$). Bottom: at the end of the stream
($R_{8,j}$). NumGLUE-cm accounts for most of the difference between the two
merges.}
\label{fig:overall}
\end{figure*}

\begin{figure*}[!t]
\centering
\includegraphics[width=\textwidth]{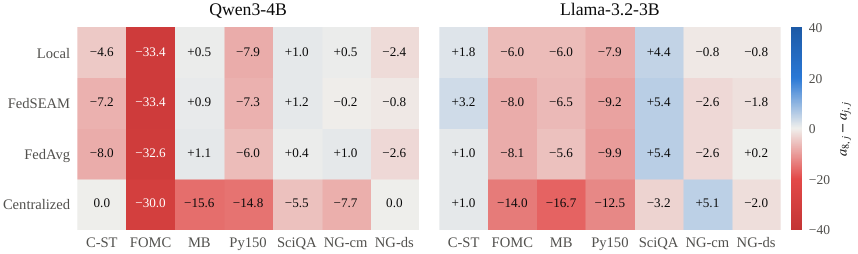}
\caption{Change of each task from right after learning it to the end of the
stream ($R_{8,j}-R_{j,j}$, 500 samples per task, equal split, client mean),
tasks 1--7. Red is
forgetting, blue is a gain.}
\label{fig:forget_heat}
\end{figure*}

With 500 samples per task, Tables~\ref{tab:pertask_qwen} and~\ref{tab:pertask_llama}
give the score of each task right after it is learned ($R_{j,j}$) and at the
end of the stream ($R_{8,j}$) on the equal split. Table~\ref{tab:nong} repeats the metrics
of Table~\ref{tab:main500} without NumGLUE-cm. Figure~\ref{fig:overall} shows the gain
of each merge over local training, and Figure~\ref{fig:forget_heat} how much each
task changes between the two points.

\begin{table*}[!t]
\caption{Per-task scores, Qwen3-4B, 500 samples per task, equal split, client
mean.}
\label{tab:pertask_qwen}
\centering\small
\setlength{\tabcolsep}{4.5pt}
\begin{adjustbox}{max width=\textwidth}
\begin{tabular}{@{}ll*{8}{c}@{}}
\toprule
Method & Row & C-STANCE & FOMC & MeetingBank & Py150 & ScienceQA & NumGLUE-cm & NumGLUE-ds & 20Minuten \\
\midrule
Local & $R_{j,j}$ & 60.80 & 66.60 & 22.88 & 42.44 & 83.60 & 43.20 & 48.00 & 37.18 \\
      & $R_{8,j}$ & 56.20 & 33.20 & 23.39 & 34.56 & 84.60 & 43.72 & 45.60 & 37.18 \\\addlinespace
FedAvg over tiles & $R_{j,j}$ & 62.40 & 69.00 & 21.76 & 41.00 & 83.60 & 49.23 & 46.80 & 37.25 \\
      & $R_{8,j}$ & 54.40 & 36.40 & 22.84 & 34.98 & 84.00 & 50.25 & 44.20 & 37.25 \\\addlinespace
\method{} & $R_{j,j}$ & 63.20 & 67.40 & 22.36 & 41.56 & 83.60 & 42.57 & 46.60 & 36.99 \\
      & $R_{8,j}$ & 56.00 & 34.00 & 23.28 & 34.27 & 84.80 & 42.41 & 45.80 & 36.99 \\\addlinespace
\textcolor{gray}{Centralized} & $R_{j,j}$ & 66.00 & 72.00 & 42.96 & 48.39 & 83.00 & 57.69 & 53.00 & 38.96 \\
      & $R_{8,j}$ & 66.00 & 42.00 & 27.37 & 33.62 & 77.55 & 50.00 & 53.00 & 38.96 \\
\bottomrule
\end{tabular}
\end{adjustbox}

\end{table*}

\begin{table*}[!t]
\caption{Per-task scores, Llama-3.2-3B-Instruct, 500 samples per task, equal
split, client mean.}
\label{tab:pertask_llama}
\centering\small
\setlength{\tabcolsep}{4.5pt}
\begin{adjustbox}{max width=\textwidth}
\begin{tabular}{@{}ll*{8}{c}@{}}
\toprule
Method & Row & C-STANCE & FOMC & MeetingBank & Py150 & ScienceQA & NumGLUE-cm & NumGLUE-ds & 20Minuten \\
\midrule
Local & $R_{j,j}$ & 51.00 & 64.60 & 39.69 & 45.52 & 81.40 & 42.82 & 44.00 & 39.56 \\
      & $R_{8,j}$ & 52.80 & 58.60 & 33.68 & 37.58 & 85.78 & 42.05 & 43.20 & 39.56 \\\addlinespace
FedAvg over tiles & $R_{j,j}$ & 50.20 & 67.20 & 39.99 & 46.76 & 81.60 & 52.31 & 44.60 & 39.59 \\
      & $R_{8,j}$ & 51.20 & 59.11 & 34.40 & 36.84 & 86.98 & 49.74 & 44.80 & 39.59 \\\addlinespace
\method{} & $R_{j,j}$ & 49.20 & 65.60 & 39.70 & 46.49 & 80.80 & 44.87 & 44.60 & 39.53 \\
      & $R_{8,j}$ & 52.40 & 57.60 & 33.17 & 37.32 & 86.15 & 42.31 & 42.80 & 39.53 \\\addlinespace
\method{}+R & $R_{j,j}$ & 50.00 & 66.80 & 39.76 & 47.11 & 81.20 & 47.69 & 45.40 & 39.52 \\
      & $R_{8,j}$ & 50.80 & 58.40 & 33.33 & 37.82 & 86.15 & 43.85 & 44.20 & 39.52 \\\addlinespace
\textcolor{gray}{Centralized} & $R_{j,j}$ & 58.00 & 70.00 & 41.39 & 41.64 & 84.00 & 60.26 & 52.00 & 39.64 \\
      & $R_{8,j}$ & 59.00 & 56.00 & 24.69 & 29.14 & 80.81 & 65.38 & 50.00 & 39.64 \\
\bottomrule
\end{tabular}
\end{adjustbox}

\end{table*}

\begin{table}[htbp]
\caption{Metrics without NumGLUE-cm (500 samples per task; LA and OP over seven
tasks, BWT over six). \method{}+R is reported for Llama-3.2-3B only.}
\label{tab:nong}
\centering\small
\setlength{\tabcolsep}{4pt}
\begin{adjustbox}{max width=\columnwidth}
\begin{tabular}{@{}lccc|ccc@{}}
\toprule
& \multicolumn{3}{c|}{Qwen3-4B} & \multicolumn{3}{c}{Llama-3.2-3B} \\
Method & LA & OP & BWT & LA & OP & BWT \\
\midrule
\multicolumn{7}{@{}l}{\textit{Equal split}} \\
Local & 51.64 & 44.96 & $-$7.80 & 52.25 & 50.17 & $-$2.43 \\
FedAvg & 51.69 & 44.87 & $-$7.96 & 52.85 & 50.42 & $-$2.84 \\
\method{} & 51.67 & 45.02 & $-$7.76 & 52.27 & 49.85 & $-$2.83 \\
\method{}+R & -- & -- & -- & 52.83 & 50.03 & $-$3.26 \\
\textcolor{gray}{Centr.} & \textcolor{gray}{57.76} & \textcolor{gray}{48.36} & \textcolor{gray}{$-$10.97} & \textcolor{gray}{55.24} & \textcolor{gray}{48.47} & \textcolor{gray}{$-$7.90} \\
\bottomrule
\end{tabular}
\end{adjustbox}

\end{table}

\subsection{Per-Client Results}\label{app:clients}
\begin{figure}[htbp]
\centering
\includegraphics[width=\columnwidth]{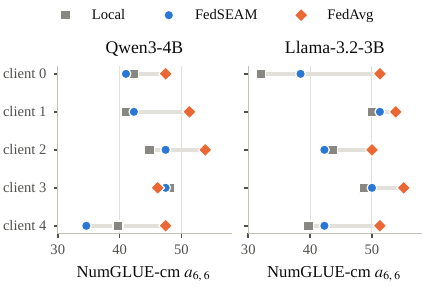}
\caption{NumGLUE-cm score of each client right after the task is learned (equal
split).}
\label{fig:clients}
\end{figure}

\begin{figure*}[htbp]
\centering
\includegraphics[width=0.8\textwidth]{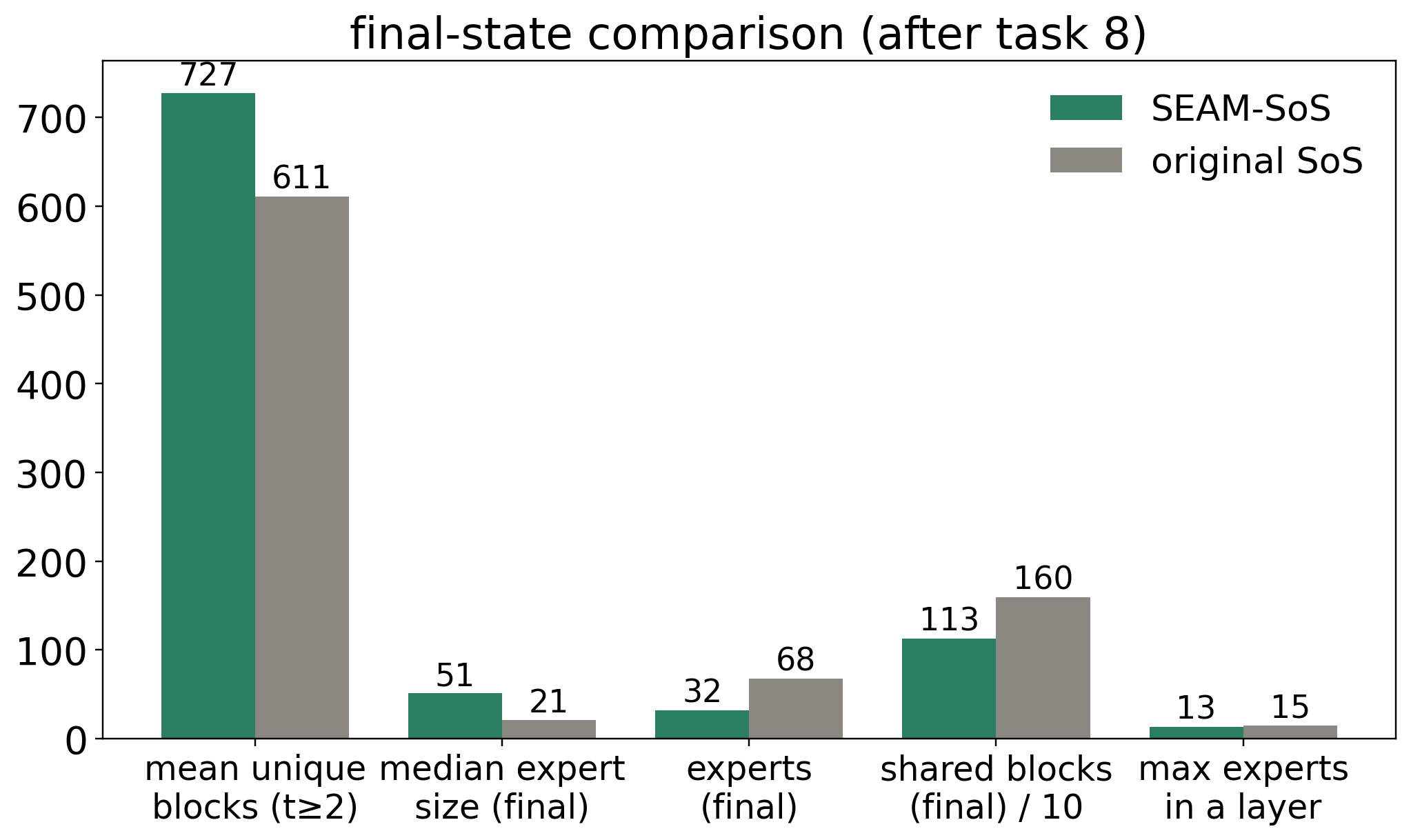}
\caption{LA and OP of every client's model (small markers) and the mean of each
method (large markers; the Local mean is drawn hollow because it lies almost on
the \method{} mean). The star is the centralized model.}
\label{fig:client_scatter}
\end{figure*}

Table~\ref{tab:perclient} gives LA, OP and BWT for each client's model.
Table~\ref{tab:ngcm_clients} and Figure~\ref{fig:clients} give the NumGLUE-cm
score of each client right after the task is learned, and
Figure~\ref{fig:client_scatter} places every client's model by its LA and OP.

\begin{table*}[!t]
\caption{LA / OP / BWT of each client's model (500 samples per task).}
\label{tab:perclient}
\centering\small
\setlength{\tabcolsep}{3.5pt}
\begin{adjustbox}{max width=\textwidth}
\begin{tabular}{@{}ll*{5}{c}@{}}
\toprule
Setting & Method & client 0 & client 1 & client 2 & client 3 & client 4 \\
\midrule
\multirow{3}{*}{Qwen3-4B, equal} & Local & 50.46 / 41.86 / $-$9.83 & 50.55 / 47.20 / $-$3.82 & 51.26 / 44.14 / $-$8.13 & 50.90 / 46.52 / $-$5.00 & 49.79 / 44.31 / $-$6.26 \\
 & FedAvg & 50.97 / 46.14 / $-$5.51 & 52.16 / 46.11 / $-$6.92 & 51.96 / 44.32 / $-$8.74 & 50.88 / 46.27 / $-$5.26 & 50.94 / 44.85 / $-$6.95 \\
 & \method{} & 50.06 / 42.91 / $-$8.17 & 50.96 / 46.78 / $-$4.78 & 51.80 / 44.14 / $-$8.76 & 50.79 / 43.68 / $-$8.13 & 49.06 / 45.97 / $-$3.54 \\
\midrule
\multirow{4}{*}{Llama-3.2-3B, equal} & Local & 49.51 / 48.38 / $-$1.30 & 52.90 / 50.21 / $-$3.08 & 50.62 / 49.30 / $-$1.51 & 51.33 / 49.29 / $-$2.32 & 51.02 / 48.60 / $-$2.77 \\
 & FedAvg & 53.02 / 49.99 / $-$3.46 & 53.23 / 50.65 / $-$2.95 & 53.00 / 50.50 / $-$2.86 & 52.05 / 50.38 / $-$1.91 & 52.59 / 50.14 / $-$2.80 \\
 & \method{} & 50.02 / 47.71 / $-$2.64 & 52.60 / 49.84 / $-$3.16 & 51.41 / 50.14 / $-$1.46 & 51.47 / 48.96 / $-$2.87 & 51.24 / 47.91 / $-$3.81 \\
 & \method{}+R & 51.32 / 48.99 / $-$2.66 & 53.47 / 50.08 / $-$3.88 & 51.81 / 49.94 / $-$2.13 & 52.26 / 48.42 / $-$4.38 & 52.07 / 48.86 / $-$3.68 \\
\bottomrule
\end{tabular}
\end{adjustbox}

\end{table*}

\begin{table}[htbp]
\caption{NumGLUE-cm score of each client right after the task ($R_{6,6}$, equal
split). }
\label{tab:ngcm_clients}
\centering\small
\setlength{\tabcolsep}{3.5pt}
\begin{adjustbox}{max width=\columnwidth}
\begin{tabular}{@{}ll*{5}{c}@{}}
\toprule
& Method & c0 & c1 & c2 & c3 & c4 \\
\midrule
\multirow{3}{*}{Qwen} & Local & 42.31 & 41.03 & 44.87 & 48.05 & 39.74 \\
 & FedAvg & 47.44 & 51.28 & 53.85 & 46.15 & 47.44 \\
 & \method{} & 41.03 & 42.31 & 47.44 & 47.44 & 34.62 \\
\midrule
\multirow{4}{*}{Llama} & Local & 32.05 & 50.00 & 43.59 & 48.72 & 39.74 \\
 & FedAvg & 51.28 & 53.85 & 50.00 & 55.13 & 51.28 \\
 & \method{} & 38.46 & 51.28 & 42.31 & 50.00 & 42.31 \\
 & \method{}+R & 43.59 & 53.85 & 43.59 & 52.56 & 44.87 \\
\bottomrule
\end{tabular}
\end{adjustbox}

\end{table}


\subsection{Server Aggregation in Detail}\label{app:server}
\begin{figure*}[!t]
\centering
\includegraphics[width=\textwidth]{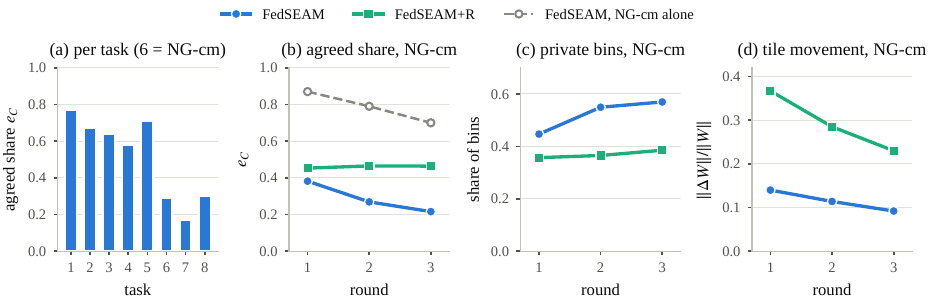}
\caption{Server aggregation on Llama-3.2-3B. (a) Share of a client's update that
lies in the agreed subspace, $e_C$, mean over the three rounds of each task.
(b--d) The rounds of NumGLUE-cm (task 6): (b) agreed share, (c) share of bins
left private, (d) median change the merge makes to a client's tile, relative to
the tile. \method{}+R also averages the part of the update outside the agreed
subspace.}
\label{fig:server}
\end{figure*}

For each tile coordinate held by at least two clients, the server builds the
common subspace $\mathcal{C}$ from the clients' update subspaces (energy $0.9$,
at most $32$ directions). $e_C$ is the share of a client's update energy inside
$\mathcal{C}$, and a bin is private when nothing is agreed or $e_C<0.5$. Tile
movement is the median, over accepted tiles, of the size of the change divided by
the size of the tile. On Llama-3.2-3B, \method{} merges $58$--$77\%$ of each
update on the first five tasks ($e_C$ of $0.77$, $0.67$, $0.64$, $0.58$ and
$0.71$) but only $17$--$30\%$ on the last three ($0.29$, $0.17$ and $0.30$).
Figure~\ref{fig:server} and Table~\ref{tab:server_t6} follow the rounds of
NumGLUE-cm. Trained as a single task, the clients agree on $87\%$, $79\%$ and
$70\%$ of their updates in the three rounds.

\begin{table}[htbp]
\caption{Server statistics for NumGLUE-cm (task 6), Llama-3.2-3B. Accepted tiles:
range over the five clients of accepted tiles over all tiles of the task.}
\label{tab:server_t6}
\centering\small
\setlength{\tabcolsep}{2.6pt}
\begin{adjustbox}{max width=\columnwidth}
\begin{tabular}{@{}lcccccc@{}}
\toprule
& \multicolumn{3}{c}{\method{}} & \multicolumn{3}{c}{\method{}+R} \\
Round & 1 & 2 & 3 & 1 & 2 & 3 \\
\midrule
Bins & 1942 & 1942 & 1942 & 1946 & 1946 & 1946 \\
Private bins & 868 & 1067 & 1106 & 695 & 711 & 750 \\
$e_C$ & 0.382 & 0.269 & 0.216 & 0.453 & 0.465 & 0.464 \\
$\lambda$ & 0.896 & 0.907 & 0.896 & 0.890 & 0.924 & 0.935 \\
Sign-test drops & 7038 & 8180 & 9288 & 6199 & 4527 & 4402 \\
Accepted tiles & 71--82\% & 72--83\% & 71--82\% & 72--82\% & 72--83\% & 72--83\% \\
Tile movement & 0.140 & 0.114 & 0.092 & 0.367 & 0.285 & 0.230 \\
\bottomrule
\end{tabular}
\end{adjustbox}

\end{table}

\subsection{Additional Ablations}\label{app:ablation}
Table~\ref{tab:abl_task} gives the per-task scores of the protection ablations of
Table~\ref{tab:ablation}, Table~\ref{tab:abl_more} repeats them on other client
subsets, Table~\ref{tab:abl_ref} changes the reference point of the client
update, and Table~\ref{tab:sweep} varies the settings on a four-task stream.

\begin{table*}[!t]
\caption{Per-task scores of the protection ablations, Llama-3.2-3B, clients 0--3.}
\label{tab:abl_task}
\centering\small
\setlength{\tabcolsep}{4.5pt}
\begin{adjustbox}{max width=\textwidth}
\begin{tabular}{@{}ll*{8}{c}@{}}
\toprule
Configuration & Row & C-STANCE & FOMC & MeetingBank & Py150 & ScienceQA & NumGLUE-cm & NumGLUE-ds & 20Minuten \\
\midrule
no protection & $R_{j,j}$ & 49.25 & 65.50 & 37.00 & 43.56 & 81.00 & 42.31 & 45.25 & 39.48 \\
 & $R_{8,j}$ & 50.25 & 60.75 & 32.11 & 38.60 & 87.00 & 37.50 & 44.25 & 39.48 \\\addlinespace
gate freezing only & $R_{j,j}$ & 49.00 & 66.00 & 36.89 & 43.41 & 80.50 & 43.27 & 45.25 & 39.49 \\
 & $R_{8,j}$ & 50.50 & 60.50 & 31.71 & 43.37 & 86.75 & 38.46 & 44.50 & 39.49 \\\addlinespace
\method{} & $R_{j,j}$ & 49.00 & 65.25 & 39.75 & 46.17 & 81.00 & 45.51 & 44.75 & 39.58 \\
 & $R_{8,j}$ & 51.50 & 57.75 & 33.08 & 37.98 & 86.47 & 41.67 & 45.25 & 39.58 \\
\bottomrule
\end{tabular}
\end{adjustbox}

\end{table*}

\begin{table}[htbp]
\caption{Protection ablations on other client subsets (Llama-3.2-3B).}
\label{tab:abl_more}
\centering\small
\setlength{\tabcolsep}{3pt}
\begin{adjustbox}{max width=\columnwidth}
\begin{tabular}{@{}lccc@{}}
\toprule
Configuration & LA & OP & BWT \\
\midrule
\multicolumn{4}{@{}l}{\textit{All five clients}} \\
\method{} & 51.35\std{0.92} & 48.91\std{1.10} & $-$2.79\std{0.86} \\
\quad no protection & 50.43\std{0.98} & 48.79\std{0.74} & $-$1.88\std{1.22} \\
\midrule
\multicolumn{4}{@{}l}{\textit{Clients 0--1}} \\
\method{} & 51.31 & 48.77 & $-$2.90 \\
\quad merge protection only & 51.19 & 48.16 & $-$3.46 \\
\quad gate freezing only & 50.42 & 49.60 & $-$0.94 \\
\quad no protection & 50.12 & 49.33 & $-$0.91 \\
\bottomrule
\end{tabular}
\end{adjustbox}

\end{table}

\begin{table}[htbp]
\caption{Reference point of the client update on the server. Measuring updates
from the start of the task instead of the start of the round changes no metric
by more than 0.2 points.}
\label{tab:abl_ref}
\centering\small
\setlength{\tabcolsep}{3pt}
\begin{adjustbox}{max width=\columnwidth}
\begin{tabular}{@{}llccc@{}}
\toprule
Backbone & Update reference & LA & OP & BWT \\
\midrule
\multirow{2}{*}{Qwen} & start of the round & 50.53\std{1.15} & 44.76\std{1.94} & $-$6.60\std{2.45} \\
 & start of the task & 50.54\std{1.03} & 44.69\std{1.62} & $-$6.67\std{2.35} \\
\midrule
\multirow{2}{*}{Llama} & start of the round & 51.15\std{1.13} & 48.80\std{1.08} & $-$2.69\std{1.11} \\
 & start of the task & 51.35\std{0.92} & 48.91\std{1.10} & $-$2.79\std{0.86} \\
\bottomrule
\end{tabular}
\end{adjustbox}

\end{table}

\begin{table}[htbp]
\caption{Four-task stream (C-STANCE, FOMC, MeetingBank, Py150), Qwen3-4B, five
clients unless stated, $R_{j,j}$ and LA.}
\label{tab:sweep}
\centering\small
\setlength{\tabcolsep}{2.5pt}
\begin{adjustbox}{max width=\columnwidth}
\begin{tabular}{@{}lccccc@{}}
\toprule
Configuration & C-ST & FOMC & MB & Py150 & LA \\
\midrule
FedAvg over tiles & 56.00 & 66.20 & 20.41 & 40.49 & 45.78 \\
\method{} ($\theta{=}0.7$) & 56.80 & 66.00 & 20.35 & 41.48 & 46.16 \\
\quad conflict threshold 0 & 56.60 & 66.00 & 20.32 & 41.92 & 46.21 \\
\quad conflict threshold $-0.3$ & 56.60 & 66.00 & 20.47 & 41.18 & 46.06 \\
\quad $\theta=0.8$ & 56.60 & 65.80 & 20.49 & 41.26 & 46.04 \\
\quad no divergence term & 56.60 & 66.80 & 20.62 & 41.75 & 46.44 \\
\quad absolute layer floor & 59.00 & 67.60 & 21.22 & 42.12 & 47.49 \\
\quad labels written back & 59.40 & 67.40 & 21.38 & 43.39 & 47.89 \\
\quad budget 12\% & 55.20 & 66.60 & 20.01 & 40.79 & 45.65 \\
\quad budget 25\% & 63.20 & 67.80 & 21.43 & 45.36 & 49.45 \\
\quad one round per task & 53.20 & 66.40 & 8.25 & 19.27 & 36.78 \\
\quad 10 clients & 55.60 & 63.60 & 7.81 & 19.54 & 36.64 \\
FedAvg, 10 clients & 56.00 & 65.80 & 7.84 & 19.76 & 37.35 \\
\bottomrule
\end{tabular}
\end{adjustbox}

\end{table}

\subsection{Expert Structure along the Task Stream}\label{app:structure}
The pipeline can be stopped after the split of every task, which yields the
expert structure of all tasks within hours; experts are untrained in this
protocol, so the direction test compares gradients with gradients.
Tables~\ref{tab:sos_summary} and~\ref{tab:sos_structure} compare the direction-aware split with sharing
by position in such a structure-only pass, with one client holding all data
(Qwen3-4B, budget $12\%$, expansion limit $1.5$, and an earlier task order). The
gain in tiles of the task's own expert is positive from the third task on and largest at Py150 ($+314$). 
\begin{table*}[htbp]
\caption{Expert structure along the task stream. Own is the number of tiles in the task's own expert,
Experts the number of experts present after the task, and Shared the number of tiles held by shared
experts. Tasks are listed in the order used in this pass.}
\label{tab:sos_structure}
\centering
\small
\begin{tabular}{ll rrr rrr r}
\toprule
& & \multicolumn{3}{c}{direction-aware} & \multicolumn{3}{c}{positional} & \\
\cmidrule(lr){3-5}\cmidrule(lr){6-8}
$t$ & task & Own & Experts & Shared & Own & Experts & Shared & $\Delta$Own \\
\midrule
1 & C-STANCE   & 800  & 1  & 0    & 800 & 1  & 0    & $0$    \\
2 & FOMC       & 473  & 3  & 351  & 500 & 3  & 327  & $-27$  \\
3 & MeetingBank& 869  & 4  & 351  & 715 & 7  & 404  & $+154$ \\
4 & ScienceQA  & 649  & 9  & 657  & 601 & 13 & 746  & $+48$  \\
5 & NumGLUE-cm & 753  & 12 & 657  & 614 & 21 & 860  & $+139$ \\
6 & 20Minuten  & 720  & 19 & 892  & 604 & 30 & 1173 & $+116$ \\
7 & Py150      & 1003 & 24 & 892  & 689 & 48 & 1322 & $+314$ \\
8 & NumGLUE-ds & 624  & 32 & 1127 & 555 & 68 & 1596 & $+69$  \\
\bottomrule
\end{tabular}

\end{table*}
\begin{table}[htbp]
\caption{Expert structure under the direction-aware split and sharing by position (structure-only
pass, one client holding all data, Qwen3-4B, budget 12\%).}
\label{tab:sos_summary}
\centering
\small
\begin{tabular}{@{}lrr@{}}
\toprule
After the eighth task & Direction & Position \\
\midrule
Mean tiles in own expert ($t{\ge}2$) & \textbf{727} & 611 \\
Experts present            & \textbf{32} & 68 \\
Median expert size (tiles) & \textbf{51} & 21 \\
Tiles in shared experts    & \textbf{1127} & 1596 \\
Total tiles held           & 2933 & 2940 \\
\bottomrule
\end{tabular}
\end{table}
\end{document}